\documentclass{ieeeaccess}
\usepackage{cite}
\usepackage{amsmath,amssymb,amsfonts}
\usepackage{physics, bm}
\usepackage{algorithm}
\usepackage{algpseudocode}
\usepackage{graphicx}
\usepackage{textcomp}
\usepackage{float}
\usepackage{subcaption}
\usepackage{comment}

\usepackage{siunitx}
\newcommand{\vect}[1]{\vb{#1}}          
\newcommand{\cDV}{\Delta V}             
\newcommand{\pmiss}{d_{\text{pm}}}      
\newcommand{\vclose}{v_{\text{close}}}  

\usepackage[utf8]{inputenc}
\usepackage{booktabs}
\usepackage{enumitem}
\usepackage{graphicx}
\usepackage{array}
\usepackage{longtable}

\usepackage[hyphens]{url}
\usepackage[hidelinks]{hyperref}

\usepackage{tikz}
\usetikzlibrary{
  calc,
  angles,
  quotes,
  arrows.meta,
  decorations.markings,
  positioning,
  fit
}

\usepackage{tikz}
\usetikzlibrary{positioning,calc,arrows.meta}%
\usetikzlibrary{angles,quotes,decorations.markings,fit,intersections}
\usepackage{pgfplots}
\pgfplotsset{compat=newest}

\tikzset{
  body/.style={circle,draw,thick},
  force/.style={-Latex,very thick},
  orbit/.style={thick,blue},
  debris/.style={circle,fill=black,inner sep=0.8pt},
  good/.style={thick,green!60!black},
  bad/.style={thick,red},
  hint/.style={dashed,gray!60},
  mass/.style={inner sep = 2.2pt, circle},
  relvec/.style={-latex, blue, thick},
  absvec/.style={-latex, thick},
  safe/.style={dashed, gray},
  debrisdot/.style={fill=black, circle, inner sep=1pt}
}

\newcommand{\pcord}[3]{\coordinate (#3) at ({#1*cos(#2)},{#1*sin(#2)});}

\NewSpotColorSpace{PANTONE}
\AddSpotColor{PANTONE} {PANTONE3015C} {PANTONE\SpotSpace 3015\SpotSpace C} {1 0.3 0 0.2}
\SetPageColorSpace{PANTONE}%

\def\BibTeX{{\rm B\kern-.05em{\sc i\kern-.025em b}\kern-.08em
    T\kern-.1667em\lower.7ex\hbox{E}\kern-.125emX}}
\begin{document}
\history{Received 30 November 2025, accepted 13 January 2026, date of publication 19 January 2026, date of current version 5 February 2026.}
\doi{10.1109/ACCESS.2026.3655237}

\title{Satellite Trajectory Optimization via Proximal Policy Optimization for Space Debris Avoidance}

\author{\uppercase{Logan Luna}\authorrefmark{1}\authorrefmark{4},
\IEEEmembership{Member, IEEE},
\uppercase{Juan Ortiz Couder}\authorrefmark{2},
\IEEEmembership{Member, IEEE},
and \uppercase{Raul Alejandro Vargas-Acosta}\authorrefmark{3},
\IEEEmembership{Member, IEEE}}

\address[1]{Department of Electrical Engineering and Computer Science,
Embry-Riddle Aeronautical University, Daytona Beach, FL 32114 USA}

\address[2]{Department of Electrical Engineering and Computer Science,
Embry-Riddle Aeronautical University, Daytona Beach, FL 32114 USA
(e-mail: ORTIZCOJ@my.erau.edu)}

\address[3]{Department of Electrical Engineering and Computer Science,
Embry-Riddle Aeronautical University, Daytona Beach, FL 32114 USA
(e-mail: VARGASAR@erau.edu)}

\address[4]{Present address: School of Computer Science, College of Computing,
Georgia Institute of Technology, Atlanta, GA 30332 USA
(e-mail: lluna@gatech.edu)}

\markboth
{L. Luna \headeretal: Satellite Trajectory Optimization via PPO for Space Debris Avoidance}
{L. Luna \headeretal: Satellite Trajectory Optimization via PPO for Space Debris Avoidance}

\corresp{Corresponding author: Logan Luna (e-mail: lluna@gatech.edu).}

\begin{abstract}

Collision avoidance systems are commonly used to avoid fragmentation events occuring in Low-Earth Orbit (LEO) and Geosynchronous Equatorial Orbit (GEO). However, these events have been growing in frequency as orbital congestion worsens with the launch of megaconstellations. Consequently, conjunction alerts and collision risks are becoming increasingly common. Current practices, which are commonly manual or rule-based, have difficulty scaling to these worsening dynamic environments. To address this intensifying situation, we propose a reinforcement-learning policy for autonomous collision avoidance, trained via Proximal Policy Optimization (PPO) along with an open-source, high-fidelity astrodynamics simulator for training and evaluation. In 1{,}000 deterministic GEO episodes, our agent achieves a 97.5\% collision avoidance success rate, outperforming traditional controllers such as a rule-based baseline (20.7\% success) and an impulsive $\Delta v$ planner baseline (27.5\% success). To achieve these results, we designed a simulator to train and evaluate our agent, using real-world and simulated debris. We simulate Newtonian two-body dynamics using Sun/Moon third-body perturbations, fuel-dependent thrust, and configurable debris fields. The agent is trained with curriculum learning and shaped rewards oriented toward encouraging survival, adequate projected miss distance, and $\Delta v$ conservation. Finally, our evaluation consisted of a fully deterministic pipeline, including shared seeds, per-episode logs, and telemetry exports \footnote{Our work is a publicly available framework at \href{https://purl.org/sat-trajectory-avoidance}{https://purl.org/sat-trajectory-avoidance}}.


\end{abstract}


\titlepgskip=-15pt

\maketitle


\section{Introduction: }
\label{sec1}

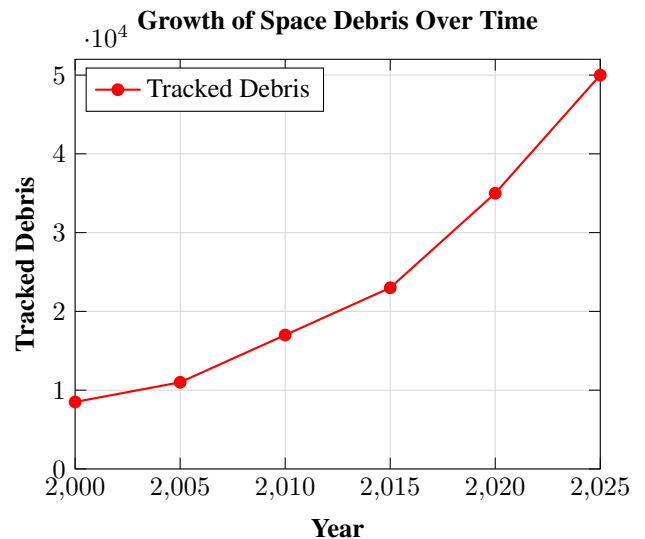
\begin{figure}[H]
    \centering
    \begin{tikzpicture}
    \begin{axis}[
        width=1\linewidth,
        height=7cm,
        title={\textbf{Growth of Space Debris Over Time}},
        xlabel={\textbf{Year}},
        ylabel={\textbf{Tracked Debris}},
        xmin=2000, xmax=2025,
        ymin=0, ymax=52000,
        xtick={2000,2005,2010,2015,2020,2025},
        grid=both,
        grid style={gray!30},
        legend style={
            at={(0.02,0.98)},
            anchor=north west,
            fill=white,
            draw=black
        },
        tick label style={color=black},
        title style={color=black},
        label style={color=black},
        axis line style={black},
        tick style={black},
    ]
    \addplot[
        color=red,
        mark=*,
        mark options={fill=red},
        thick
    ] coordinates {
        (2000, 8500)
        (2005, 11000)
        (2010, 17000)
        (2015, 23000)
        (2020, 35000)
        (2025, 50000)
    };
    \addlegendentry{Tracked Debris}
    \end{axis}
    \end{tikzpicture}
    \caption{Growth of tracked space debris from 2000 to 2025. Values are shown in units of $10^4$ (i.e., the tick value 1 represents 10,000 pieces).}
    \label{fig:space_debris}
\end{figure}

The Earth's orbital environment is becoming congested at a near exponential rate according to the European Space Agency \cite{esa2024} (Fig. \ref{fig:space_debris}). Currently there are over 40,500 tracked objects larger than 10\,cm, 1.1 million untracked objects between 1–10\,cm and 130 million between 1\,mm–1\,cm \cite{esa2024}. Due to this increasingly cluttered environment, their annual fragmentation events are projected to increase owing to collisions and satellite failures. This is further accelerated by satellite launches driven by megaconstellations and commercial expansion, with a projected 28,000+ new satellites projected by 2033 \cite{PayloadSpace2024}. Additionally, a small subset of all satellites launched by humans are still functional as of today, with the ESA's Space Debris Office estimating that only half of satellites in space are still operational in 2021 \cite{esa2021numbers}.

Without proactive mitigation, this increasing debris density threatens orbital sustainability and raises collision probabilities, accelerating the Kessler Syndrome~\cite{Kessler1978}, a feedback loop of cascading collisions. While the existing literature and agencies have explored procedural or manual collision avoidance, these systems are generally non-autonomous, proprietary, and lack long-term orbital considerations. Therefore, these systems are often insufficient or infeasible for dynamic, debris-rich environments.

We aim to address the need for intelligent and sustainable handling of these cascading issues by developing a machine learning (ML) driven autonomous collision avoidance system for geostationary satellites. Additionally, we aimed to create a publicly available astrodynamics simulation environment for training, testing, and monitoring based on current dynamically changing debris environments. Using this system, we train a reinforcement learning (RL) Proximal Policy Optimization (PPO) agent that balances collision avoidance, orbital stability, and fuel conservation.

The contributions of this study include the following. Realistic orbital simulator with Newtonian gravity, third-body perturbations, and fuel-aware thrust modeling. A Gym-compatible RL environment optimizes safety, efficiency, and orbital accuracy. Open-source training and evaluation pipeline for debris-rich Geosynchronous Equatorial Orbit (GEO) scenarios. Our simulated results show reduced collisions, improved stability, and lower fuel use compared to baselines.

The broader impact of our study includes supporting intelligent space-flight operations, space traffic management, operational resilience, and long-term orbital sustainability.

\section{Background}

\subsection{Orbital Congestion}
Earth’s orbits, particularly Low Earth Orbit (LEO), are becoming increasingly congested with satellites and debris. In 2023 alone, more than 2,800 new satellites were launched into LEO \cite{PayloadSpace2024}. Most of these satellites are concentrated at altitudes between 500 and 600 km, creating a dense orbital band where nearly two-thirds of all active satellites currently reside \cite{PayloadSpace2024}. This clustering significantly increases collision risks, since the average LEO satellite now receives approximately 30 close conjunction alerts each year \cite{PayloadSpace2024}. 


Each close approach also introduces the possibility of fragmentation, which would generate additional debris and further increase the probability of future collisions—a phenomenon known as the Kessler Syndrome \cite{Kessler1978}. Managing these risks requires international coordination, advanced tracking capabilities, and more efficient space traffic management systems. Without proactive measures, the rapid expansion of satellite constellations and continued debris accumulation threaten the safe and sustainable use of LEO for communications, Earth observation, and scientific missions.

\subsection{Space Debris Growth}
At the end of 2024, more than 50,000 tracked objects larger than 10\,cm are in orbit \cite{BGR2024}, a sharp increase from one year earlier when it was approximately 35,000 \cite{PayloadSpace2024}, even with proposed rules for commercial launches requiring upper stages to be disposed of in one of five ways (controlled reentry, transfer to graveyard orbit, Earth escape orbit, active removal within five years, or uncontrolled atmospheric disposal under certain conditions) \cite{FAA2023}. This rapid growth reflects the continuing accumulation of space debris from defunct satellites, spent rocket stages, and fragmentation events. Beyond cataloged debris, estimates suggest that 1.2 million pieces larger than 1\,cm and hundreds of millions of smaller fragments also exist in orbit \cite{BGR2024, UNU2024}. Even subcentimeter debris, traveling at orbital speeds of 7 to 8 km/s, has the kinetic energy to puncture shielding or damage critical spacecraft components.

Ongoing fragmentation events further worsen the situation. According to the European Space Agency, roughly ten such events occur every year, each injecting new debris into already congested orbital regions \cite{SpaceIntelDebris2024}. For example, the 2024 explosion of a 6A Long March rocket stage alone created more than 700 new fragments \cite{BGR2024}. This accumulation increases operational risks for active satellites and accelerates the likelihood of a runaway Kessler Syndrome scenario \cite{Johnson2010OrbitalDebris, Tarran2021PrepareForImpact}. Without effective mitigation measures, experts warn that debris collisions could, within a few decades, surpass natural satellite failures as the primary source of new orbital debris, threatening the long-term sustainability of space activities.

\subsection{Megaconstellations and Future Growth}
The rise of megaconstellations is expected to multiply the number of active satellites in orbit in the coming years. Multiple companies and agencies have proposed adding broadband constellations of satellites that would increase by tens of thousands the number of active satellites. Even under conservative forecasts, it is projected that there will be 58,000 new satellites by 2030 \cite{GAO2023}, while more aggressive forecasts predict more than 100,000 satellites by 2030 \cite{UNU2024b}. For context, the global satellite population in 2022 was only about 8,000 \cite{UNU2024c}, underscoring the scale of this expansion.

This growth rate is already alarming the space community. The European Space Agency's 2024 report explicitly warns that the current deployment strategies are "unsustainable" unless new mitigation measures are adopted \cite{PayloadSpace2024}. In some key Low Earth Orbit altitudes, the number of active satellites now rivals the amount of tracked debris \cite{BGR2024}, creating conditions of extreme congestion. This convergence of large constellations and preexisting debris fields amplifies the risks of collision and long-term orbital instability. To manage these challenges, experts stress the urgent need for advanced, automated collision-avoidance systems and comprehensive space traffic management frameworks that can coordinate satellite operations at scale.

\subsection{Current Satellite Collision Avoidance Practices}

\subsubsection{Surveillance and Alerts}
Satellite collision avoidance today relies heavily on space surveillance networks, with the U.S. Space Surveillance Network (SSN) serving as the primary global operator. The SSN continuously tracks cataloged objects in orbit, predicts close approaches, and distributes standardized Conjunction Data Messages (CDMs) \cite{AI4EarthScienceCDMTransparency}. A typical LEO satellite may receive hundreds of CDMs per week, although most are automatically screened and filtered out as low-risk encounters \cite{ESA2025CDM}. After this filtering process is performed, operators are left with approximately two actionable alerts per satellite per week, which require manual assessment, and in some cases, the planning of collision-avoidance maneuvers \cite{ESA2025CDM}. These practices illustrate the growing operational burden posed by the expanding satellite population.

\subsubsection{Human-in-the-Loop Decision Process}
Collision avoidance in current satellite operations relies on a human-in-the-loop decision framework, where analysts evaluate predicted conjunctions and determine whether a maneuver is necessary. Commonly adopted probability thresholds guide these decisions: for uncrewed satellites, a collision probability of \(10^{-4}\) for uncrewed satellites, while crewed missions apply a more conservative threshold of \(10^{-5}\) \cite{ScienceDirectCAMThreshold, NatureThreshold}. For instance, the ESA typically issues a collision avoidance command if the predicted chance of collision exceeds \(1/10{,}000\). When this command is triggered, Collision Avoidance Maneuvers (CAM) involve small orbital adjustments executed hours or days before the predicted conjunction to ensure sufficient separation and reduce the risk of impact. This human-centric approach balances operational caution with practical resource management but can become increasingly challenging as the number of conjunctions rises with the growth of active satellites.

\subsubsection{Manual Coordination and Communication}
In addition to automated alerts and human-in-the-loop assessment, satellite operators must often engage in direct coordination to manage potential collisions. Communication is often conducted via email as no formalized “right-of-way” rules exist to assign maneuver responsibility \cite{SpacePolicyOnline2019}. This informal process can create operational vulnerabilities, as demonstrated by a 2019 near-miss between ESA's Aeolus satellite and SpaceX's Starlink. In this instance, ESA was required to perform the CAM because the lack of timely communication prevented the coordinator of the Starlink operator \cite{SpacePolicyOnline2019}. Such events highlight the limitations of current manual coordination processes and the growing need for standardized communication protocols and automated space traffic management systems as orbital congestion intensifies.

\subsubsection{Maneuver Planning}
Collision avoidance maneuvers are carefully planned using flight dynamics software, which calculates optimal \(\Delta v\) burns to ensure sufficient separation while minimizing fuel consumption and mission disruption \cite{ArXivCAMPlanning}. Various approaches are employed in this planning process, including analytical solutions, numerical simulations, and heuristic optimization techniques. Despite these tools, significant human supervision is still required as operators must define mission constraints, validate proposed maneuvers, and ensure that the solutions align with broader operational objectives. This combination of automated computation and expert oversight allows satellites to safely execute collision avoidance maneuvers while preserving limited propellant and maintaining mission performance.

\subsubsection{Timing and Execution}
The timing of collision avoidance maneuvers is critical, as CDMs are continually updated to reflect evolving orbital information. Operators often postpone decision-making to incorporate the latest data, reducing the likelihood of false alarms and unnecessary maneuvers. However, this approach frequently compresses the window for final analysis, maneuver approval, and command upload to just a few hours before the predicted conjunction. As a result, operators must balance the benefits of up-to-date data against the operational risks of limited execution time.

\subsubsection{Current Effectiveness}
Historically, collision avoidance practices have been sufficient to manage the relatively small population of active satellites. For example, the European Space Agency averages roughly one collision avoidance maneuver per satellite per year \cite{ESA2025CDM}, while NASA and the International Space Station conduct occasional orbital adjustments, typically a few burns annually \cite{SpaceStackExchangeISS}. These statistics showcase that, until recently, manual and semi-automated methods were largely adequate to mitigate collision risks. However, as the number of satellites and debris continues to surge, these traditional practices are increasingly insufficient, underscoring the need for more automated, scalable, and coordinated space traffic management solutions.

\subsection{Limitations of Manual and Ad Hoc Methods}

\subsubsection{Limited Autonomy}
Current collision avoidance practices rely heavily on human judgment and manual processes, which are increasingly strained as satellite populations grow \cite{ArcivAutonomyStrain2024}. With thousands of conjunctions occurring in Low Earth Orbit each week, these manual methods are not scalable, creating the potential for delayed responses or missed collision warnings. Recognizing these limitations, officials at the European Space Agency have called for the development of formal space traffic management frameworks and greater automation in maneuver planning and execution \cite{SpacePolicyOnline2019}. Such enhancements would enable operators to handle larger volumes of conjunction data more efficiently while maintaining mission safety.

\subsubsection{Fuel and Mission Impact}
Collision avoidance maneuvers consume significant fuel, directly reducing satellite operational lifespan. The use of fixed probability thresholds, such as \(10^{-4}\) for uncrewed satellites, create suboptimal trade-offs: if the threshold is set too low, it will result in frequent, and often unnecessary maneuvers, yet, if the threshold is set too high, the collision risk is elevated. For example, Starlink initially used a conservative threshold of \(10^{-5}\), which resulted in frequent maneuvers. This threshold was later switched to \(10^{-6}\), resulting in roughly 50,000 avoidance maneuvers over six months (\textasciitilde14 per satellite) \cite{SpaceComFuelImpact2024}. While conservative thresholds enhance safety, they also accelerate propellant depletion and limit the effective lifetime of spacecraft \cite{SpaceComFuelImpact2024}. Current collision avoidance methods lack comprehensive global optimization that balances fuel efficiency with mission objectives.

\subsubsection{Scalability and Coordination Challenges}
Human-led collision avoidance systems are increasingly unable to manage the complexity of megaconstellations, where thousands of active satellites operate simultaneously. Starlink’s onboard automation provides a notable exception, allowing some level of autonomous maneuvering \cite{SpaceComFuelImpact2024}. However, the absence of standardized inter-operator coordination creates a "traffic log" in which the effects of maneuvers are difficult to track across different operators. Avoidance maneuvers can alter orbits by up to 40 \,km, complicating the prediction and confusing the tracking systems \cite{NatureCounterManeuver2024}. When this happens, and multiple autonomous systems act independently, without coordinating with each other, the risk of counter-maneuvers increases, thus causing more counter-maneuvers as a result iteratively \cite{NatureCounterManeuver2024}.

\subsubsection{Transparency and Data Sharing}
A significant limitation of current collision avoidance practices is the lack of transparency and widespread data sharing among satellite operators. Many operators treat maneuver data as proprietary, which restricts public access to datasets that are essential for research, model development, and independent verification. Exceptions to this norm are rare, such as the ESA's 2019 anonymized CDM dataset \cite{AI4EarthScienceCDMTransparency}. Some progress to improving the transparency has been made in the recent years, like China's publication of daily bulletins for its space station, or SpaceX's sharing Starlink ephemerides publicly since 2021 \cite{NatureTransparency2024}. Nevertheless, international experts continue to call for broader transparency measures, including shared maneuver thresholds, best-practice norms, and more accessible data to improve the safety and sustainability of orbital operations \cite{NatureTransparency2024}.

\subsubsection{Conclusion}
The current collision avoidance practices highlights that existing systems are not scalable for the rapid growth in orbital population. There is an urgent need for increased automation, intelligent decision-making, and globally coordinated space traffic management. To ensure the long-term safety and sustainability of orbital operations, a transition away from ad hoc, human-centric responses  towards a more systematic, AI-driven collision avoidance framework must be made. This transition will allow collision frameworks to handle large volumes or data while minimizing operational risks.

\subsection{Machine Learning}
Machine Learning (ML) is a subfield of artificial intelligence (AI) that focuses on designing algorithms and systems that can learn patterns from some data and improve their performance over time without being explicitly programmed to do so \cite{Alpaydin2021}. There are three main types of ML based on the training methods: Supervised Learning, Unsupervised Learning, and Reinforcement Learning (RL). In Supervised Learning, the model learns from labeled data in the form of input-output pairs, and models its behavior to match underlying trends within that data \cite{alloghani2020}. For Unsupervised learning, the model aims to find structured to unlabeled data, without knowing the output to that data, and clusters the given data into groups based on their behavior \cite{alloghani2020}. Last, RL makes the model learn by interacting with the environment, upon each interaction, the model receives either a rewards if the model's behavior is correct, or a penalty if the behavior is incorrect \cite{franccois2018}. 

\begin{figure}
    \centering
    \includegraphics[width=1\linewidth]{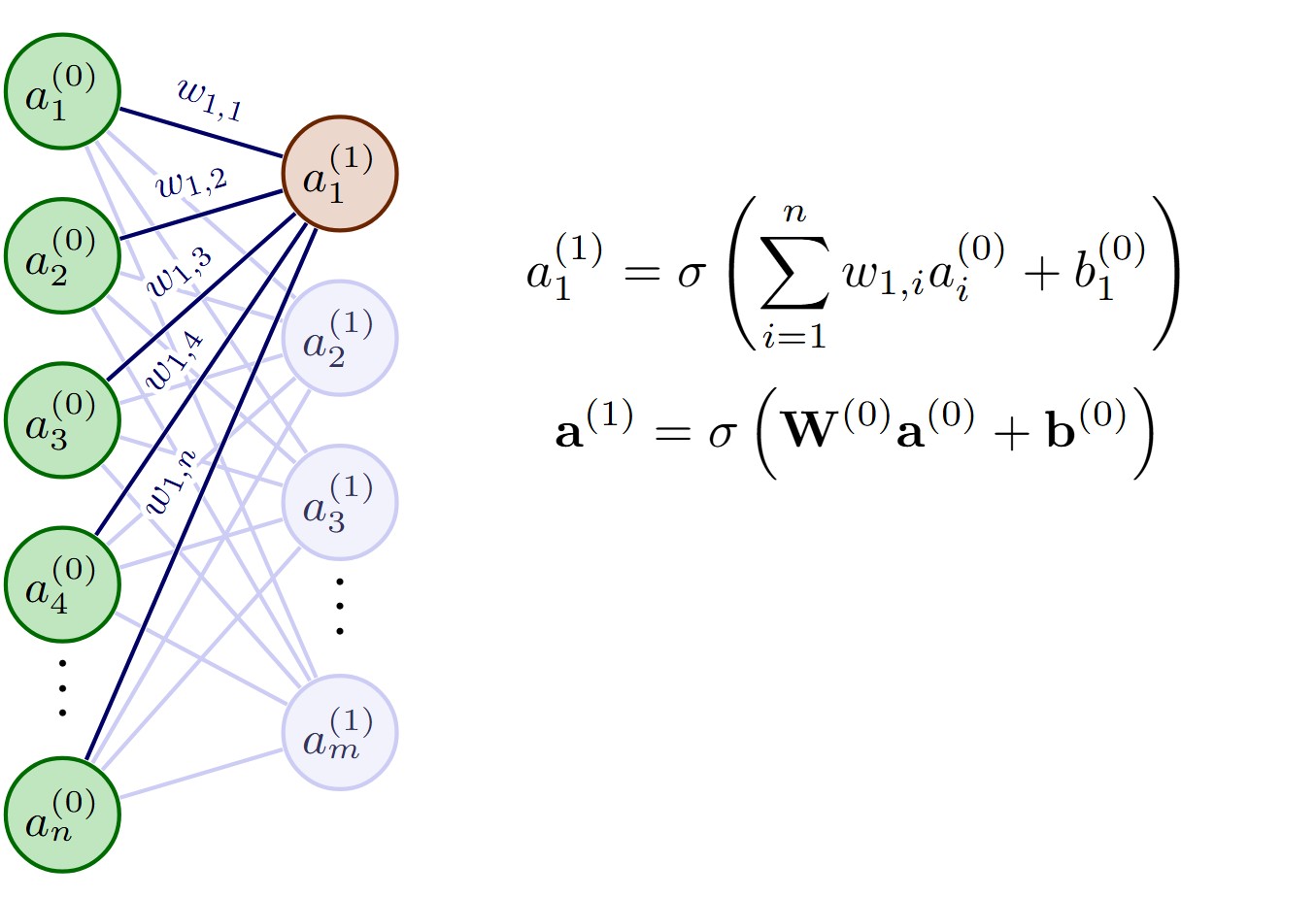}
    \caption{A single layer’s forward pass: input vector $\mathbf{a}^{(0)}$ transforms into $\mathbf{a}^{(1)}$ via weights $w_{j,i}$, biases $b_j$, and an activation $\sigma(\cdot)$.}
    \label{fig:activation}
\end{figure}

Fig. \ref{fig:activation} shows an architecture of a ML model. It shows two layers that are filly connected, also known as dense layers, which are the most common within ML models. Every single neuron is connected to every neuron of the next layer. Each neuron receives an input from every neuron in the previous layer, and the importance of each input is known as its weight. 

\subsection{Reinforcement Learning}
Reinforcement Learning (RL) is a paradigm in machine learning in which an agent learns to make decisions by interacting with an environment to maximize cumulative rewards.

\begin{figure}
    \centering
    \includegraphics[width=\linewidth]{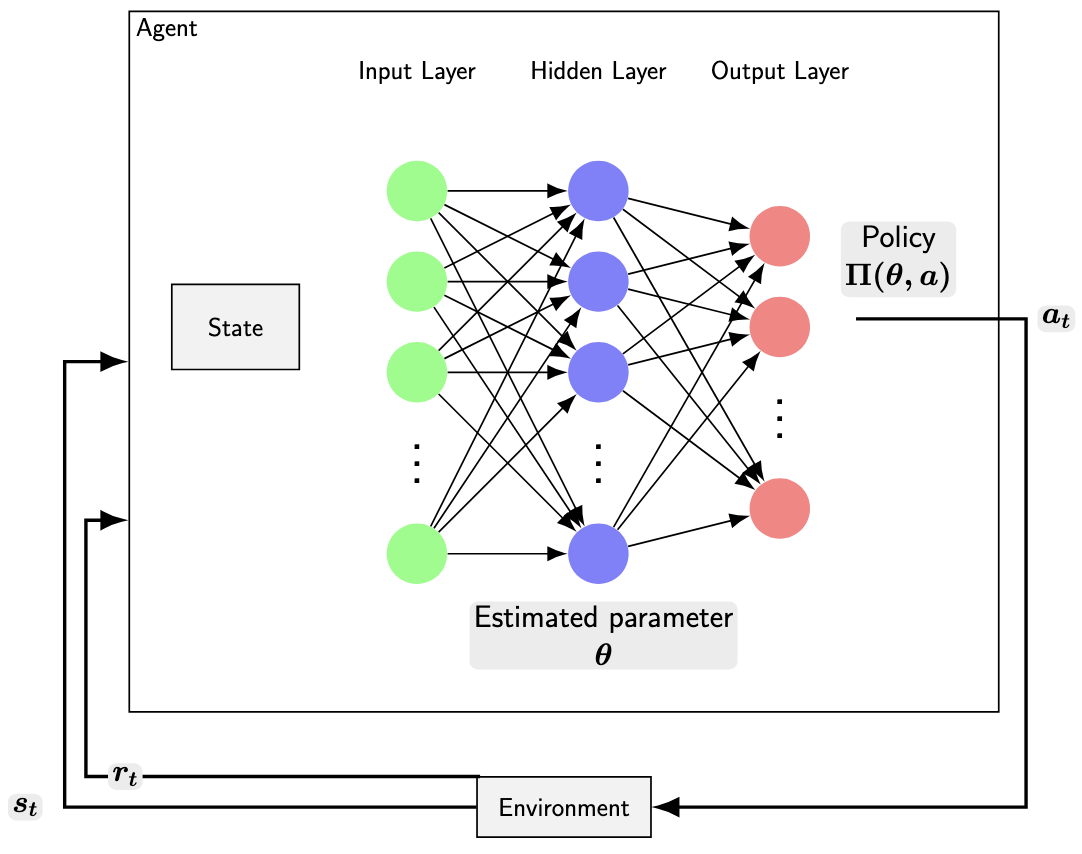}
\caption{
Neural network architecture illustrating agent–environment interaction. 
$s_t$ is the environment state at time $t$, 
$a_t$ is the action chosen by the agent, 
$r_t$ is the reward returned by the environment, 
and $\theta$ denotes the network’s learned parameters. 
The policy $\Pi(\theta,a)$ uses $\theta$ to map states to actions.
}
\label{fig:rl_network}
\end{figure}

\subsubsection{Environment, States, and Agent}
In RL, the environment represents the system in which an agent interacts, providing states ($s_t$) that describe the current situation and rewards ($r_t$) based on the agent's actions ($a_t$). The agent then interacts with the environment based on these results, selecting actions to maximize the expected cumulative reward. This creates an interaction loop that involves the agent perceiving the state, choosing an action, and receiving feedback from the environment, as shown in Fig. \ref{fig:rl_network}.

\subsubsection{Actions and Rewards}
Actions refer to possible moves or decisions that an agent can make in each state. The choice of actions influences the next state and the rewards received, which guides the agent towards desirable behaviors. The objective of the agent is to learn a policy $\Pi$ that maps states to actions, maximizing the long term reward.

\subsection{Proximal Policy Optimization (PPO)}
Proximal Policy Optimization (PPO) is a popular reinforcement learning algorithm designed to provide stable and efficient policy updates \cite{schulman2017ppo}. These updates leads to balanced exploration, stability, and sample efficiency, resulting in it being commonly used for continuous-control tasks. PPO is a policy gradient method that optimizes behavior through adjusting the policy parameters with the aim of increasing the expected cumulative reward. In order to promote stability, PPO has a clipped surrogate objective function. This restricts the policy updates, preventing large changes to limit potentially negative behavior. In addition to this, it utilizes an actor-critic architecture, where two neural networks are employed. The actor network outputs actions based on a learned policy, and the critic network estimates the value function to guide policy improvement. This improves learning efficiency and reduces variance in gradient estimates. Lastly, PPO uses Gaussian distributions to model continuous action spaces. This makes it perform well in continuous environments that require precise, continuous action. As a result of this design we employ PPO as the primary architecture in our study.

\subsection{Baseline Controllers}
\label{sec:baselines}

In our study we evaluate our work with variations of the following baseline controllers.

\subsubsection{Impulsive $\Delta v$ Planner}
This baseline implements an avoidance strategy focused on classical orbital mechanics and conjunction analysis. This strategy applies a constant lateral acceleration, with the goal of increasing the projected miss distance beyond a defined safety threshold. 

Previous studies using methods similar to this generally compute a thrust ($\Delta v$) to maximize miss distance. This is done by using the relative position and velocity between the primary and secondary objects \cite{armellin2021collision}. Additionally, an operational conjunction assessment is used to apply spatial and probabilistic thresholds to decide when to execute avoidance burns \cite{chan2008spacecraft}. The controller also uses linearized relative motion models for the propagation and maneuver design \cite{battin1999astrodynamics}. As a result, these impulsive methods yield predictable, interpretable trajectories suitable as benchmarks for learning-based systems \cite{armellin2021collision, chan2008spacecraft}.

\subsubsection{Rule-Based Controller}
The rule-based baseline approaches collision avoidance by evaluating relative kinematics for all debris fragments and calculating the thrust magnitude based on a risk metric. 

This controller computes relative position $\vect{r}_{\text{rel}}$ and velocity $\vect{v}_{\text{rel}}$ for estimating the closing speed to the debris in order to determine the best avoidance action \cite{battin1999astrodynamics}. This controller is commonly used by agencies such as NASA and ESA, which employ assessment frameworks that trigger avoidance maneuvers when spatial or temporal thresholds are violated \cite{nasa2015cara,esa2021sdo}. While this is a more classical approach, recent studies have proposed combining spatial proximity and time pressure into scalar risk indices, using them to guide the controllers maneuver magnitude and direction \cite{park2019decision,hwang2020heuristic,ardaens2013collision}. Overall, rule-based controllers provide predictable responses while serving as a common baselines for evaluating learning-based collision avoidance policies.

\subsubsection{Deep Q-Network (DQN)}
In addition to the traditional controllers used in collision avoidance, we also include a baseline reinforcement learning policy to compare the performance of our introduced PPO policy. This policy is a Deep Q-Network (DQN) that learns as state-action value function via a neural approximator updated by temporal-difference targets \cite{mnih2015dqn}. For a more in depth explanation regarding architecture and policy design, please refer to \ref{sec:DQN_Method}.

\subsubsection{No-Action Controller}
This baseline serves as an evaluation tool as a sanity check, verifying if the environment correctly detects collisions. In our application it represents a passive satellite that performs no collision-avoidance maneuvers. As a result, it reflects the inherent risk of simulated scenarios and demonstrates the effectiveness of collision-avoidance systems.

\section{Related Work}
\subsection{Toward Autonomous and ML-Driven Collision Avoidance}
Manual collision avoidance, due to requiring manual handling, are recognized as slow, reactive, and unsustainable given the rapid growth of orbital populations. In response, some space agencies and researchers are developing automated systems designed to minimize human involvement in conjunction handling. The overarching goal is to enable satellites, or their ground-based software counterparts, to autonomously assess collision risk, plan optimal maneuvers, and execute these rapidly and consistently. As an example, ESA is prototyping a fully automated collision avoidance system that autonomously manages the entire workflow, from risk assessment, to maneuver execution, and leverages the space-based communication links for real-time command uploads \cite{ESAAuto2024}. Similarly, U.S. Space Policy Directive-3 (SPD-3), advocates for enhanced Space Traffic Management (STM) tools to support more efficient and scalable operations \cite{SPD3_2018}. These developments reflect a broader vision to transition from reactive, manual approaches, to a more streamlined, autonomous collision avoidance capable of safely handling the growing density of satellites in orbit.

\subsubsection{Machine Learning for Conjunction Analysis and Decision Support}
ML offers significant advantages for collision avoidance by providing adaptability and data-driven decision-making beyond fixed-rule approaches. Early applications of ML have focused on improving conjunction analysis and prioritizing risk among other alerts. For example, ESA's Collision Avoidance Challenge invited participants to develop ML models to predict whether initial alerts would escalate into serious collision threats \cite{ESAChallenge2021}. The resulting \textit{Kessler} library is an open-source Bayesian deep learning tool that leverages recurrent neural networks to forecast the evolution of conjunctions \cite{KesslerLibrary2021,KesslerLibrary2022}. By identifying high-risk events more accurately, the tool helps operator focus their attention where it is most needed, reducing cognitive load. Similarly, NASA's Conjunction Assessment Risk Analysis (CARA) program explored AI and ML classifiers for automated "Go/No-Go" maneuver decisions \cite{NASA_CARA2019}, emphasizing front-end filtering and decision support rather than full autonomy. ML-based risk analysis enhance early-stage decision-making, aiding operators by providing actionable insights, yet, they stop short of fully replacing human oversight. 

\subsubsection{Reinforcement Learning for Maneuver Planning and Execution}
RL provides a promising approach for enabling autonomous spacecraft to learn optimal collision avoidance maneuvers through trial-and-error interactions within simulated environments. RL is particularly well-suited for sequential decision-making under uncertainty, where trade-offs between safety, fuel consumption, and mission objectives must be continuously balanced. Bourriez et al.\ (2023) framed the collision avoidance problem as a partially observable Markov decision process (POMDP), training an AI agent to autonomously decide CAMs without human intervention \cite{POMDP2023}. The primary objective is to delegate decision-making to spacecraft itself, enabling faster, decentralized responses that don't rely on ground-based commands. This represents a significant shift from traditional ground-controlled operations towards onboard AI autonomy, potentially allowing satellites to respond more rapidly and efficiently to the growing number of satellites in orbital environments.

\subsubsection{Proximal Policy Optimization (PPO) Applications}
Proximal Policy Optimization (PPO) has emerged as a state-of-the-art deep reinforcement learning algorithm, valued for its stability, robustness, and efficiency in high-dimensional decision spaces \cite{schulman2017ppo}. Recent research demonstrates its strong potential for autonomous collision avoidance in LEO. Kazemi et al.\ (2024) applied PPO to train a LEO satellite for optimal avoidance planning over an eight-day horizon, achieving significantly lower collision risk and reduced fuel consumption compared to heuristic methods \cite{Kazemi2024}. While their implementation does not appear to be publicly available, our work emphasizes full transparency to support reproducibility. Similarly, Smith et al.\ (2024) report that PPO-based policies consistently outperform rule-based baselines in maneuver decision-making \cite{Smith2024}. The core advantage of PPO lies in its ability to dynamically adapt to evolving orbital conditions and to identify non-intuitive maneuvers that go beyond the scope of human-designed rules.

\subsubsection{Openness and Transparency in ML-Driven Approaches}
A recurring theme in the development of ML-driven collision avoidance systems is the emphasis on openness and reproducibility. Many initiatives adopt open-source frameworks and leverage open simulators such as Python-based environments or OpenAI's Gym, to encourage community validation and rapid iteration \cite{OpenAI_Gym2016,StableBaselines2017}. ESA's \textit{Kessler} library exemplifies this approach by providing an openly available Bayesian deep learning tool for conjunction forecasting \cite{KesslerLibrary2021}. The openness in this domain carries several benefits, including the promotion of peer review and reproducibility, increased trust in autonomous decision-making systems, and the facilitation of cross-operator adoption for unified avoidance strategies. In contrast, many industrial implementations of ML remain proprietary and opaque, limiting transparency and slowing collaboratively practices.

\subsubsection{Summary and Future Directions}
ML-driven methods offer faster reaction times, improved fuel efficiency, and scalability for large constellations. Open, collaborative frameworks strengthen this transition by improving transparency and trust among different stakeholders. Nevertheless, several important challenges still remain, including the rigorous validation of RL policies for safety-critical missions and the development of coordination mechanisms to prevent conflicting maneuvers between independently operated agents. The convergence of advanced algorithms, high-fidelity data, and open collaboration points towards a transformative shift from manual responses to a more proactive and autonomous collision avoidance, supported by data-driven space traffic management practices and sustainable orbital operations \cite{ESAAuto2024,SPD3_2018}.

\section{Methodology}
In the following sections, we discuss the creation of our simulation environment along with the training and evaluation of our proposed PPO model. 

\subsection{Astrodynamic Satellite Simulation}
\begin{figure}
    \centering
    \includegraphics[width=0.5\textwidth]{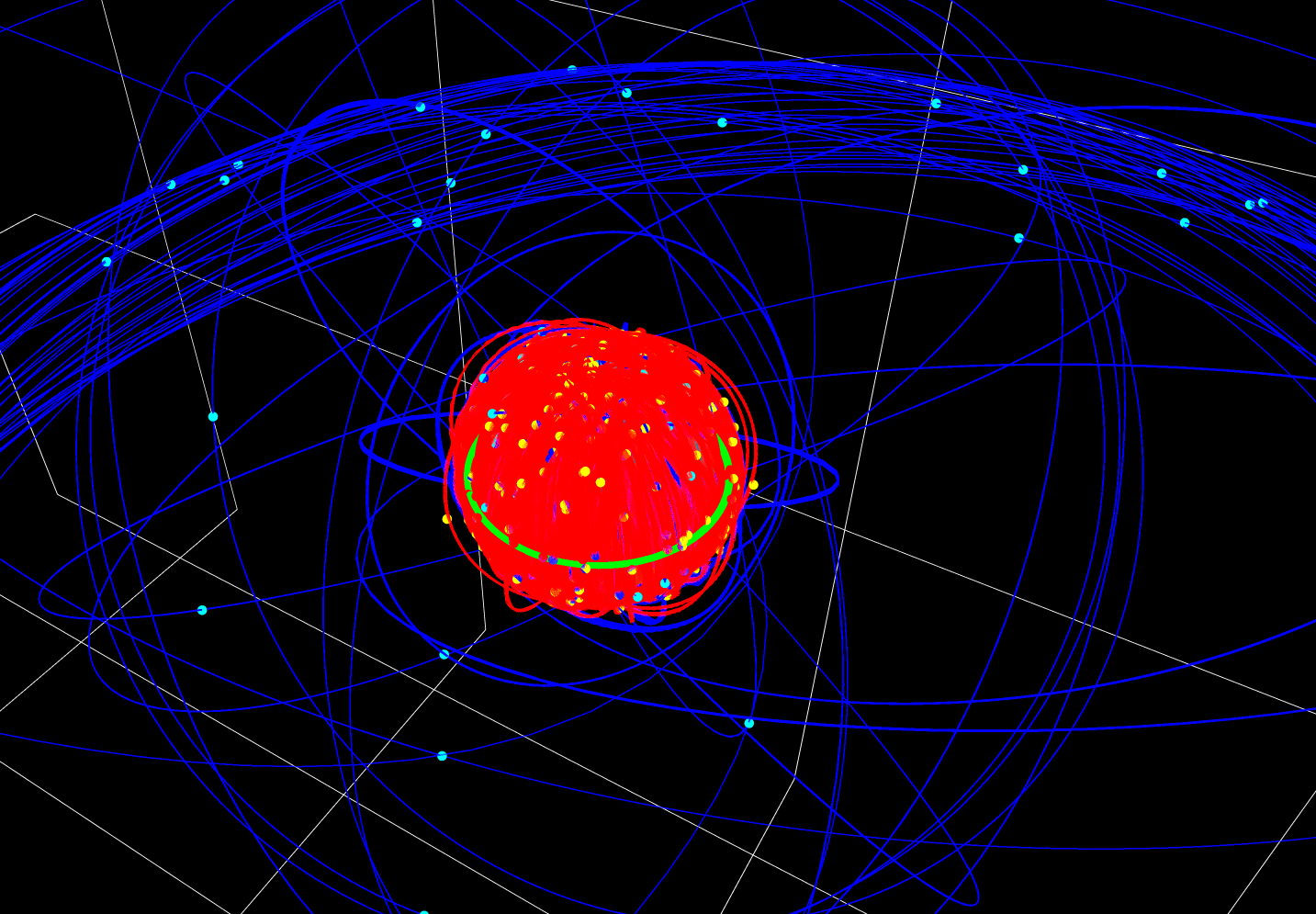}
    \caption{Model Orbit 
    \\ Yellow Dots: Debris Red Lines: Debris Trajectories 
    \\ Blue Dots: Active Satellites Blue Lines: Active Satellites Trajectories 
    \\ Green Dot: Agent Green Line: Agent Trajectory}
    \label{fig:model-orbit}
\end{figure}

Below, we describe the creation steps and usage of the environments in our simulation. Fig. \ref{fig:model-orbit} shows a sample in our simulated environment. 

\subsubsection{Initialization}

\begin{figure*}[t]
\centering
\begin{tikzpicture}[line cap=round,scale=1]
  \coordinate (Earth) at (0,0);
  \coordinate (Moon)  at (6,1.2);
  \coordinate (Sun)   at (-7,2.5);

  \node[body,fill=blue!45,draw=blue!70,minimum size=16pt] at (Earth) {};
  \node[below=6pt] at (Earth) {\small Earth}; 

  \node[body,fill=gray!50,draw=gray!70,minimum size=8pt] at (Moon) {};
  \node[below] at (Moon) {\small Moon};

  \node[body,fill=yellow!70!orange,draw=orange!80!black,minimum size=18pt] at (Sun) {};
  \node[above=6pt] at (Sun) {\small Sun}; 

  \def\ro{3.3}   
  \def\ry{2.1}   
  \def\ang{35}
  \draw[orbit,draw=green!60!black] (2.6,0)
        arc[start angle=0,end angle=360,x radius=\ro,y radius=\ry];

  \coordinate (Sat) at ({-2.6 + \ro*cos(\ang)},{\ry*sin(\ang) + 0.82}); 
  \node[debris] at (Sat) {};
  \fill (Sat) circle [radius=1.7pt, draw=green!50!black, fill=green!70!black]; 
  \node[above right] at (Sat) {\small Satellite};

  \draw[force,blue!80]  (Sat) -- ($(Sat)!0.8!(Earth)$) node[midway,below right] {\(\vec F_{\!E}\)};
  \draw[force,gray!70]  (Sat) -- ($(Sat)!0.85!(Moon)$)  node[midway,above]       {\(\vec F_{\!M}\)};
  \draw[force,orange!90!black] (Sat) -- ($(Sat)!0.92!(Sun)$) node[midway,above left] {\(\vec F_{\!\odot}\)};

\end{tikzpicture}
\caption{Third-body perturbations in an Earth-centered frame. Gravitational contributions from the Moon and Sun act alongside Earth’s gravity on the satellite; the simulator sums these accelerations to produce the total force model used for propagation and control. Vectors not to scale.}
\label{fig:third_body}
\end{figure*}
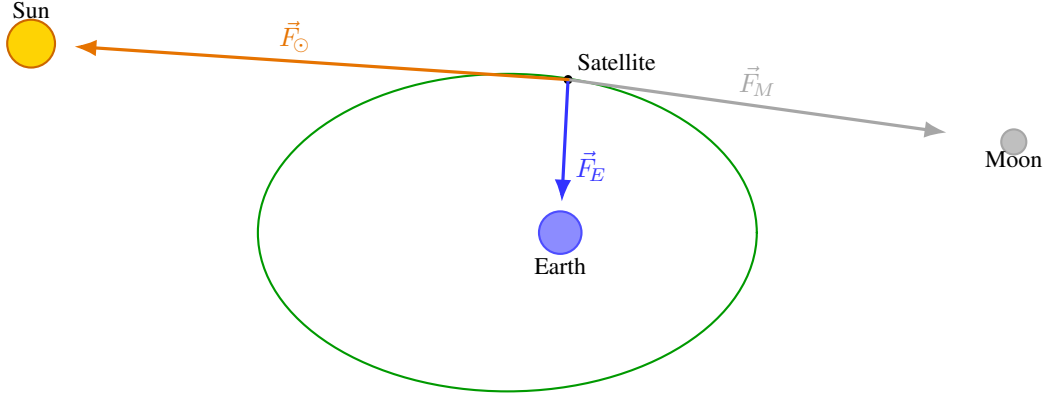

To initialize our environment, depending on the user's selection, the simulation either randomly generate debris and active satellites in LEO or pull real-time, active locations. If real-time locations are used, publicly available orbital debris state vectors and Two-Line Element (TLE) entries are retrieved from the CelesTrak database \cite{CelesTrakTLE2024}. After collecting and reformatting this data, we convert each TLE entry into an orbital state vectors using standard simplified perturbations model propagation to initialize the satellite orbits. Subsequently, we define the simulation duration and sample time steps across this period for trajectory propagation. We perform the trajectory propagation using astrodynamics primitives from the libraries \textit{Astropy} \cite{astropy2022} and \textit{Poliastro} \cite{poliastro2019}. In each episode we calculate the propagation of an active satellite subjected to cumulative thrust force, gravity from the Earth, Moon, and Sun, and optional dynamic debris velocities. Visualizations of this are shown in Figs. \ref{fig:third_body}, \ref{fig:kepler_law}, and \ref{fig:rel_geometry}. The configurable parameters of our simulation are presented in Appendix \ref{configurations}. To not overfit on real-world data, we generate a simulated dataset for debris positions for training. Using a set seed the simulation generates states from configured ranges, including number of debris, initial relative positions/velocities, and time-to-collision (TTC). These ranges are based on trends in real data.

\subsubsection{Physical Dynamics Modeling}

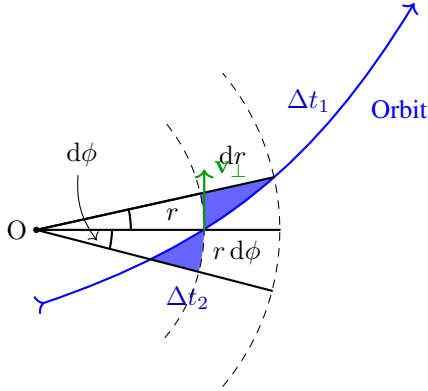
\begin{figure}
\centering
\begin{tikzpicture}[line cap=round]
  \def\r{2.22}
  \def\rr{3.22}

  \coordinate (O) at (0,0);
  \pcord{\rr}{12.5}{A}
  \pcord{\r}{12.5}{B}
  \pcord{\rr}{-14.5}{C}
  \pcord{\r}{-14.5}{D}
  \pcord{1.54}{-14.5}{E}
  \coordinate (F) at (\r,0);

  \draw[shift={(0,-1)}, thick, blue, >->] (0,0) to[bend right=20] (5,4)
      node[shift={(-0.2,-1.4)}] {Orbit};

  \node at (2.6,1) {$\dd r$};
  \node at (1.8,0.2) {$r$};
  \node at (2.65,-0.3) {$r\,\dd\phi$};
  \node (a) at (0.6,1) {$\dd\phi$};

  \path[fill=blue, opacity=0.6] (F) to[bend right=5] (B) -- (A) to[bend left=6] cycle;
  \path[fill=blue, opacity=0.6] (F) to[bend left=5] (D) -- (E) to[bend right=4] cycle;
  \node[blue!70!black] at (3.6,1.7) {$\Delta t_1$};
  \node[blue!70!black] at (2.0,-0.9) {$\Delta t_2$};

  \draw[domain=-40:40, dashed, variable=\t, samples=100] plot ({\r*cos(\t)},{\r*sin(\t)});
  \draw[domain=-40:40, dashed, variable=\t, samples=100] plot ({\rr*cos(\t)},{\rr*sin(\t)});

  \draw[thick] (O) -- (\rr,0);
  \draw[thick] (O) -- (A);
  \draw[thick] (O) -- (B);
  \draw[thick] (O) -- (C);
  \draw[->] (a) to[bend right=20] (0.8,-0.14);

  \draw[fill=black] (0,0) circle[radius=1pt] node[left] {$\mathrm{O}$};

  \pic[draw, thick, angle radius=1cm] {angle = E--O--F};
  \pic[draw, thick, angle radius=1.25cm] {angle = F--O--B};

  \draw[->,thick,green!60!black] (F) -- ++(0,0.8) node[right] {$\vb v_\perp$};
\end{tikzpicture}
\caption{Kepler’s Second Law in our simulator. The shaded sectors illustrate that the line from the focus to the satellite sweeps out equal areas in equal times (conservation of angular momentum).}
\label{fig:kepler_law}
\end{figure}


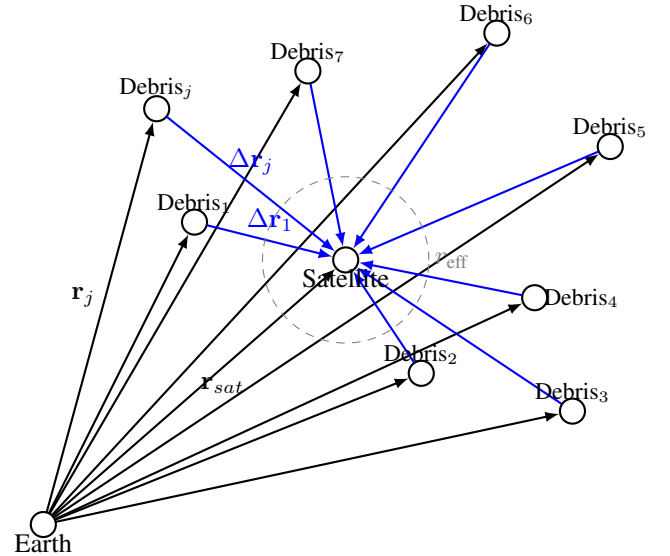
\begin{figure}
\centering
\begin{tikzpicture}[line cap=round]
  \node[body,minimum size=6pt] (O) at (0,0) {};
  \node[body,minimum size=6pt] (A) at (4,3.5) {};
  \node[body,minimum size=6pt] (B1) at (2,4) {};
  \node[body,minimum size=6pt] (B2) at (5,2) {};
  \node[body,minimum size=6pt] (B3) at (7,1.5) {};
  \node[body,minimum size=6pt] (B4) at (6.5,3) {};
  \node[body,minimum size=6pt] (B5) at (7.5,5) {};
  \node[body,minimum size=6pt] (B6) at (6,6.5) {};
  \node[body,minimum size=6pt] (B7) at (3.5,6) {};
  \node[body,minimum size=6pt] (Bj) at (1.5,5.5) {}; 

  \draw[absvec] (O) -- (A) node [pos=0.60, below] {$\vect{r}_{\rm sat}$};
  \draw[absvec] (O) -- (B1);
  \draw[absvec] (O) -- (B2);
  \draw[absvec] (O) -- (B3);
  \draw[absvec] (O) -- (B4);
  \draw[absvec] (O) -- (B5);
  \draw[absvec] (O) -- (B6);
  \draw[absvec] (O) -- (B7);
  \draw[absvec] (O) -- (Bj) node [pos=0.55, left] {$\vect{r}_{j}$};

  \draw[relvec] (B1) -- (A) node[midway,above] {$\Delta \vect{r}_1$};
  \draw[relvec] (B2) -- (A);
  \draw[relvec] (B3) -- (A);
  \draw[relvec] (B4) -- (A);
  \draw[relvec] (B5) -- (A);
  \draw[relvec] (B6) -- (A);
  \draw[relvec] (B7) -- (A);
  \draw[relvec] (Bj) -- (A) node[midway,above] {$\Delta \vect{r}_j$};

  \def\dsafe{1.1}
  \draw[safe] (A) circle (\dsafe);
  \node[gray] at ($(A)+(\dsafe+0.3,0)$) {$r_{\text{eff}}$};

  \node[below] at (O) {Earth};
  \node[below] at (A) {Satellite};
  \node[above] at (B1) {\small Debris$_1$};
  \node[above] at (B2) {\small Debris$_2$};
  \node[above] at (B3) {\small Debris$_3$};
  \node[right] at (B4) {\small Debris$_4$};
  \node[above] at (B5) {\small Debris$_5$};
  \node[above] at (B6) {\small Debris$_6$};
  \node[above] at (B7) {\small Debris$_7$};
  \node[above] at (Bj) {\small Debris$_j$};
\end{tikzpicture}
\caption{Relative geometry used for conjunction assessment. We model Earth-centric absolute states \(\vect r_{\rm sat}\) and \(\vect r_i\) and use relative separations \(\Delta \vect r_i = \vect r_i - \vect r_{\rm sat}\) (blue) to drive conjunction checks. A dashed circle of radius \(r_{\text{eff}}\) around the satellite marks the minimum separation threshold used for rewards/penalties and episode termination.}
\label{fig:rel_geometry}
\end{figure}

\begin{algorithm}
\caption{Compute Total Gravitational Acceleration}
\begin{algorithmic}[1]
\Require Satellite position $\vect{r}_{\text{sat}} \in \mathbb{R}^3$, current time $t$
\Require Constants $G$, $M_{\text{Earth}}$, $M_{\text{Moon}}$, $M_{\text{Sun}}$
\Ensure $\vect{a}_{\text{total}}$
\State $\vect{a}_{\text{Earth}} \gets -\,G\,M_{\text{Earth}}\;\dfrac{\vect{r}_{\text{sat}}}{\lVert \vect{r}_{\text{sat}} \rVert^{3}}$
\State $\vect{p}_{\text{moon}} \gets \operatorname{get\_body\_barycentric}(\texttt{'moon'},t)$; 
\State $\vect{p}_{\text{earth}} \gets \operatorname{get\_body\_barycentric}(\texttt{'earth'},t)$
\State $\vect{r}_{\text{moon,EC}} \gets \vect{p}_{\text{moon}}-\vect{p}_{\text{earth}}$ \Comment{Earth–centered moon position}
\State $\vect{r}_{\text{rel,moon}} \gets \vect{r}_{\text{moon,EC}} - \vect{r}_{\text{sat}}$,\quad
$\vect{a}_{\text{moon}} \gets G\,M_{\text{Moon}}\,\dfrac{\vect{r}_{\text{rel,moon}}}{\lVert \vect{r}_{\text{rel,moon}} \rVert^{3}}$
\State $\vect{p}_{\text{sun}} \gets \operatorname{get\_body\_barycentric}(\texttt{'sun'},t)$
\State $\vect{r}_{\text{sun,EC}} \gets \vect{p}_{\text{sun}}-\vect{p}_{\text{earth}}$
\State $\vect{r}_{\text{rel,sun}} \gets \vect{r}_{\text{sun,EC}} - \vect{r}_{\text{sat}}$,\quad
$\vect{a}_{\text{sun}} \gets G\,M_{\text{Sun}}\,\dfrac{\vect{r}_{\text{rel,sun}}}{\lVert \vect{r}_{\text{rel,sun}} \rVert^{3}}$
\State \Return $\vect{a}_{\text{total}} \gets \vect{a}_{\text{Earth}} + \vect{a}_{\text{moon}} + \vect{a}_{\text{sun}}$
\end{algorithmic}
\end{algorithm}

Orbital propagation follows Newtonian two-body dynamics with perturbations from secondary bodies. Let $\vect{r}$ and $\vect{v}$ denote the satellite position and velocity in Earth-centered inertial (ECI) coordinates and let $\vect{r}_k$ be the position of the perturbing body $k \in \{\text{Moon}, \text{Sun}\}$. \footnote{All variable definitions can be found in the nomenclature Table \ref{tab:nomenclature}.} The continuous-time equations of motion are

\begin{equation}
    \dot{\vect{r}} = \vect{v}, \qquad
\end{equation}

\begin{equation}
    \dot{\vect{v}} = -\mu_{\oplus} \frac{\vect{r}}{\|\vect{r}\|^{3}} - \sum_k \mu_k\!\left( \frac{\vect{r} - \vect{r}_k}{\|\vect{r} - \vect{r}_k\|^{3}} + \frac{\vect{r}_k}{\|\vect{r}_k\|^{3}} \right) + \frac{1}{m_t}\,\vect{T}_t
\end{equation}

where $\mu$ denotes the standard gravitational parameter, $m_t$ is the instantaneous mass, and $\vect{T}_t$ is the \textbf{thrust force} generated by the agent at time $t$. We integrate these dynamics with a semi-implicit Euler step of duration $\Delta t = 1\,\mathrm{s}$, shown in Eq. \ref{eq:thrust}:

\begin{equation}
    \vect{v}_{t+1} = \vect{v}_t + \Delta t\,\dot{\vect{v}}(\vect{r}_t, \vect{v}_t, \vect{T}_t), \quad
    \vect{r}_{t+1} = \vect{r}_t + \Delta t\,\vect{v}_{t+1}
\label{eq:thrust}
\end{equation}

Propellant expenditure follows the rocket equation with effective exhaust velocity $v_e$:

\begin{equation}
m_{t+1} = m_t - \frac{\Delta t}{v_e}\,\|\vect{T}_t\|, \qquad
\Delta v_t = \frac{\|\vect{T}_t\|}{m_t} \Delta t
\end{equation}

and we track the cumulative delta-v $\cDV_t = \sum_{i \le t} \Delta v_i$, which is fed back to the agent along with $m_t$.

Collision detection compares the instantaneous separation $d_t = \min_i\|\vect{r}_t - \vect{r}^i_{\text{debris}}\|$ to the effective safety radius $r_{\text{eff}}$. An episode terminates in collision whenever

\begin{equation}
d_t < r_{\text{eff}} = r_{\text{sat}} + r_{\text{debris}} + r_{\text{margin}}
\end{equation}

and the minimum observed $d_t$ is retained for reporting.

\begin{algorithm}
\caption{Apply Thrust to Satellite (per step)}
\begin{algorithmic}[1]
\Require Action $\vect{a}\!\in\![-1,1]^3$, $\texttt{max\_thrust}$, velocity $\vect{v}_{\text{sat}}$, masses $m_{\text{sat}},m_{\text{fuel}}$, step $\Delta t$; $I_{\text{sp}}{=}300$\,s, $g_0{=}9.80665$\,m/s$^2$
\Ensure Updated $\vect{v}_{\text{sat}}, m_{\text{sat}}, m_{\text{fuel}}, \Delta v$
\State \textbf{Sanitize action:} clamp and map to acceleration
\[
\vect{a}_{\text{cmd}} \gets \operatorname{clip}(\vect{a},-1,1),\quad
\vect{a}_{\text{thrust}} \gets \vect{a}_{\text{cmd}}\cdot \texttt{max\_thrust}
\]
\State \textbf{Integrate velocity:} apply thrust over $\Delta t$
\[
\vect{v}_{\text{sat}} \gets \vect{v}_{\text{sat}} + \vect{a}_{\text{thrust}}\,\Delta t,\qquad
\Delta v \gets \lVert \vect{a}_{\text{thrust}} \rVert\,\Delta t
\]
\State \textbf{Propellant use:} rocket equation (instantaneous approximation)
\[
m_{\text{consumed}} \gets 
\begin{cases}
m_{\text{sat}}\!\left(1 - e^{-\Delta v/(I_{\text{sp}}g_0)}\right), & \Delta v>0\\
0,& \text{otherwise}
\end{cases}
\]
\State \textbf{Mass bookkeeping:} reduce fuel and wet mass
\[
m_{\text{fuel}} \gets \max(0,\,m_{\text{fuel}}-m_{\text{consumed}}),
\]
\[
\qquad
m_{\text{sat}} \gets \,m_{\text{sat}}-m_{\text{consumed}}
\]
\State \Return $\vect{v}_{\text{sat}}, m_{\text{sat}}, m_{\text{fuel}}, \Delta v$
\end{algorithmic}
\end{algorithm}

\subsubsection{Environment Observation Space}

The observation vector for our environment concatenates the satellite position, velocity, residual fuel, and up to 100 debris positions. Additionally, this information regarding the environment is saved to a scenario blueprint, including the debris state, target index, deterministic flags, and initial fuel. This blueprint allows for the exact replication of episodic outcomes and enables trajectory visualization without rerunning the policy. Internally, the environment tracks auxiliary telemetry (elapsed time, cumulative delta-v, minimum distance to any debris, collision flags, projected miss distance) and exports these to support both post-evaluation reward shaping and analytics.

\subsection{Reinforcement Learning Environment Design}

In addition to creating our simulation software, we aim to produce a high-performing collision avoidance model. In this section, we explain the chosen model architecture, reward structure design, and model integration in our simulation.

\subsubsection{Policy Architecture and Training}

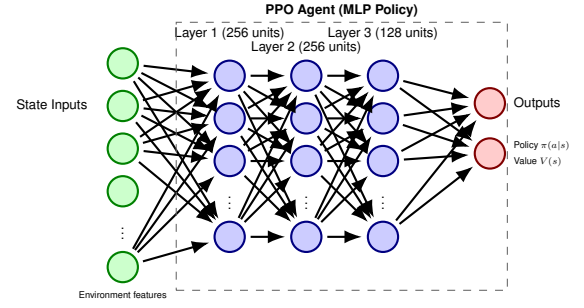
\begin{figure}
\centering
\begin{tikzpicture}[
  >=Latex,
  scale=0.5,transform shape,
  node distance=10mm and 12mm,
  every node/.style={font=\sffamily},
  layer/.style={circle, draw, thick, minimum size=8mm, inner sep=0pt},
  input/.style={layer, fill=green!20, draw=green!50!black},
  hidden/.style={layer, fill=blue!20, draw=blue!50!black},
  output/.style={layer, fill=red!20, draw=red!50!black},
  connect/.style={->, thick, shorten >=2pt, shorten <=2pt}
]
\node[input] (x1) {};
\node[input, below=3mm of x1] (x2) {};
\node[input, below=3mm of x2] (x3) {};
\node[input, below=3mm of x3] (x4) {};
\node[below=2mm of x4] (dotsin) {$\vdots$};
\node[input, below=2mm of dotsin] (xn) {};

\node[hidden, right=20mm of x2, yshift=8mm] (h1_1) {};
\node[hidden, below=3mm of h1_1] (h1_2) {};
\node[hidden, below=3mm of h1_2] (h1_3) {};
\node[below=2mm of h1_3] (dots1) {$\vdots$};
\node[hidden, below=2mm of dots1] (h1_n) {};

\node[right=of h1_1, hidden] (h2_1) {};
\node[hidden, below=3mm of h2_1] (h2_2) {};
\node[hidden, below=3mm of h2_2] (h2_3) {};
\node[below=2mm of h2_3] (dots2) {$\vdots$};
\node[hidden, below=2mm of dots2] (h2_n) {};

\node[right=of h2_1, hidden] (h3_1) {};
\node[hidden, below=3mm of h3_1] (h3_2) {};
\node[hidden, below=3mm of h3_2] (h3_3) {};
\node[below=2mm of h3_3] (dots3) {$\vdots$};
\node[hidden, below=2mm of dots3] (h3_n) {};

\node[output, right=20mm of h3_2, yshift=4mm] (out1) {};
\node[output, below=5mm of out1] (out2) {};

\foreach \i in {1,2,3,n}{
  \foreach \j in {1,2,3,n}{
    \draw[connect] (x\i) -- (h1_\j);
    \draw[connect] (h1_\i) -- (h2_\j);
    \draw[connect] (h2_\i) -- (h3_\j);
  }
}
\foreach \i in {1,2,3,n}{
  \draw[connect] (h3_\i) -- (out1);
  \draw[connect] (h3_\i) -- (out2);
}

\node[left=4mm of x2] {State Inputs};
\node[below=1mm of xn] {\scriptsize Environment features};

\node[above=4mm of h1_1] {Layer 1 (256 units)};
\node[above=0.5mm of h2_1] {Layer 2 (256 units)};
\node[above=4mm of h3_1] {Layer 3 (128 units)};

\node[right=1mm of out1] {Outputs};
\node[right=1mm of out2, align=left] {\scriptsize Policy $\pi(a|s)$ \\ \scriptsize Value $V(s)$};

\node (rightshift) [fit=(out2), xshift=-6mm] {};

\node[
    draw=black!60,
    dashed,
    fit=(h1_1) (h3_n) (rightshift),
    inner sep=5mm,
    label={[align=center, above]:\textbf{PPO Agent (MLP Policy)}}
] {};
\end{tikzpicture}
\caption{PPO agent architecture (schematic). A shared multilayer perceptron processes state inputs and produces both policy and value outputs.}
\label{fig:ppo_arch}
\end{figure}

For our agent, we chose to use a PPO architecture (shown in Fig. \ref{fig:ppo_arch}, made up of three layers, with 256, 256, and 128 nodes respectively) due to its proven reliability and performance in complex scenarios \cite{schulman2017ppo}.

PPO optimizes a clipped surrogate objective to ensure stable updates:

\begin{equation}
        L^{\mathrm{CLIP}}(\theta) =
        \hat{\mathbb{E}}_t \left[ \min\left(r_t(\theta)\,\hat{A}_t,\ \text{clip}\,(r_t(\theta),1-\epsilon,1+\epsilon)\,\hat{A}_t \right)\right]
\end{equation}

where \(r_t(\theta)\) is the probability ratio between new and old policies. Next, it combines the policy loss with a value function loss and entropy bonus for stability and exploration:
    \begin{equation}
        L^{\mathrm{PPO}}(\theta) = \hat{\mathbb{E}}_t \left[ L^{\mathrm{CLIP}}(\theta) - c_1 L^{\mathrm{VF}}(\theta) + c_2 S[\pi_\theta](s_t) \right]
    \end{equation}

Lastly, it estimates advantages using Generalized Advantage Estimation (GAE):
    \begin{equation}
        \hat{A}_t = \sum_{l=0}^{T-t-1} (\gamma \lambda)^l\,\delta_{t+l},
        \quad \delta_t = r_t + \gamma V(s_{t+1}) - V(s_t)
    \end{equation}

reducing variance while preserving learning stability.

For the implementation of this model, we use \textit{Stable-Baselines3} \cite{StableBaselines2017} with the configuration summarized in Table \ref{tab:ppo_hparams} in Appendix \ref{configurations}.

For training, we perform the curriculum shown in Table \ref{tab:curriculum}, which is enforced deterministically by the training loop, so replays of the same checkpoint experience identical scenario difficulties over time. During training, we log episodic rewards, collision flags, and delta-v usage to TensorBoard summaries and invoke deterministic evaluations every 20,000.

\begin{table*}[t]
\centering
\caption{Curriculum schedule}
\begin{tabular}{@{}lrrrr@{}}
\toprule
\textbf{Stage} &
\multicolumn{1}{c}{\textbf{Duration (steps)}} &
\multicolumn{1}{c}{\textbf{Collision Probability}} &
\multicolumn{1}{c}{\textbf{Debris Radius (m)}} &
\multicolumn{1}{c}{\textbf{Notes}} \\
\midrule
Basic avoidance    & 250{,}000 & 0.4 & 25.0 & Introduces frequent near-miss encounters. \\
Intermediate       & 350{,}000 & 0.6 & 50.0 & Forces proactive burns under moderate clutter. \\
Advanced           & 400{,}000 & 1.0 & 100.0 & Full adversarial scenarios before test regime. \\
\bottomrule
\end{tabular}
\label{tab:curriculum}
\end{table*}

\subsubsection{Reward Design and Shaping}

The shaped reward for timestep $t$ is summarized in Algorithm \ref{alg:RewardFunct}. Its features include a magnified survival and projected-miss rewards, delta-v penalties, a milestone bonus every 100 steps, and penalties for large burns and jitter to promote smooth, fuel-aware trajectories. This reward function encourages safe, fuel-efficient, and stable orbital behavior.

\begin{algorithm}
\caption{Reward \& Termination (Corrected Notation)}
\begin{algorithmic}[1]
\Require Satellite $\vect{r}_{\text{sat}}$, debris $\{\vect{r}_i\}$,
         step $\Delta v$, action change $\Delta a$, cumulative $\sum \Delta v$, step $t$
\Require Params: collision threshold $d_{\text{coll}}$, 
         safe radius $d_{\text{safe}} = d_{\text{coll}} + B$, 
         penalties/weights 
         $\{P_{\text{coll}},\lambda_{\text{surv}},c_d,\lambda_{\text{coast}},
           c_1,c_2,p_{\text{large}},\lambda_{\text{smooth}},c_{\Sigma v},
           v_{\text{soft}},\lambda_{\text{proj}},R_{\text{mile}}\}$
\Ensure $(R,\text{done})$

\State $R \gets 0$, $\text{done}\gets\text{False}$
\State $d_t \gets \min_i \| \vect{r}_{\text{sat}} - \vect{r}_i \|$ (or $+\infty$)

\If{$d_t \le d_{\text{coll}}$}
    \State \Return $(R - P_{\text{coll}},\ \text{True})$
\EndIf

\State $R \gets R + \lambda_{\text{surv}}$
        \Comment{survival reward}

\State \textbf{Distance shaping \& coasting:}
\[
R \gets R +
\begin{cases}
 c_d \cdot \min(d_t - d_{\text{safe}},\,8000)
 \; \\ +\; \mathbf{1}[\Delta v < 0.005]\,\lambda_{\text{coast}},
 & d_t > d_{\text{safe}}, \\[6pt]
 -0.00015 \cdot \min(d_{\text{safe}} - d_t,\,8000),
 & \text{otherwise.}
\end{cases}
\]

\State \textbf{Projected miss / closing speed:}
\[
R \gets R +
\lambda_{\text{proj}} \cdot
\min\!\big(\max(\pmiss - d_{\text{coll}},\,0),\,15000\big)
\]
\[
+\; 0.5 \cdot
\max\!\Big(0,\;
\frac{v_{\text{close},t-1} - v_{\text{close},t}}
     {\max(v_{\text{close},t-1}, 10^{-6})}
\Big)
\]

\State \textbf{$\Delta v$ cost (context-aware):}
\[
\text{if }\Delta v>0:\quad
C_{\Delta v} \gets (c_1\Delta v + c_2\Delta v^2)\cdot
\begin{cases}
 0.6, & d_t \le d_{\text{safe}},\\
 1, & \text{otherwise.}
\end{cases}
\]
\[
R \gets R - C_{\Delta v}
 - \mathbf{1}[\Delta v > 0.2]\,
   p_{\text{large}}(\Delta v - 0.2)^2
\]

\State \textbf{Smoothness \& cumulative cap:}
\[
R \gets R 
 -\mathbf{1}[\Delta v < 0.25]\,\lambda_{\text{smooth}}\Delta a^2
\]
\[
-\ \mathbf{1}[\sum\Delta v > v_{\text{soft}}]\,
    c_{\Sigma v}\,(\sum\Delta v - v_{\text{soft}})^{1.3}
\]

\State \textbf{Milestone reward:}
\If{$t>0$ and $t \bmod 100 = 0$}
   \State $R \gets R + R_{\text{mile}}$
\EndIf

\State \Return $(R,\text{done})$

\end{algorithmic}

\vspace{2mm}
{\footnotesize
\textit{Note:} 
Projected miss distance is $\pmiss$ (m) and relative closing speed is 
$v_{\text{close},t}$ (m/s), estimated over a short lookahead horizon.
}
\label{alg:RewardFunct}
\end{algorithm}

\subsubsection{Model and Simulation Integration}
To integrate our model with the previously described simulation, we use the \textit{OpenAI Gym} API to support autonomous collision avoidance \cite{OpenAI_Gym2016}. The environment provides an observation space that includes the agent's state, position, velocity, fuel mass, and the positions the of surrounding debris. Additionally, the environment is a continuous action space, allowing the agent to make small velocity adjustments leading to a larger change over time and requiring precise decision making. The agent's primary focus in our scenarios is to make continuous orbital control decisions aimed at avoiding collisions while maintaining stable and efficient trajectories.

\subsection{Evaluation and Testing}

\subsubsection{Baseline Controllers}

In our evaluation, we compare our trained agent against three deterministic baselines, each of which are implemented based on the general literature surrounding them, as described in Section \ref{sec:baselines}.

\paragraph{Risk-aware rule-based controller} 
These baselines uses the relative kinematics of the scenario, estimating the closing speed and time-to-closest-approach for each debris fragment. Subsequently, maneuvers are triggered when either spatial thresholds (e.g. 4$\times$ collision distance) or temporal limits (e.g. $<$240 s) are violated. The initial implementation of this baseline achieved a 0\% success rate, so to improve its performance we replaced the threshold heuristic with a multi-factor decision rule, improving the baselines performance and making it a competitive comparison. This improvement was achieved by using the following steps:

For debris body $i$, let $d_i$ be the relative distance, $\tau_i$ the time-to-closest-approach (TCA), and $\hat{\mathbf{u}}_i$ the unit vector from the satellite toward the debris. We compute a scalar risk score

\begin{equation}
\rho_i = \sigma\!\left( \alpha_d \frac{r_{\text{eff}}}{d_i} + \alpha_\tau \frac{\tau_{\text{crit}} - \tau_i}{\tau_{\text{crit}}} \right),
\end{equation}

where $\sigma$ is the logistic function, $\tau_{\text{crit}} = 240\,\mathrm{s}$, and $(\alpha_d, \alpha_\tau) = (3.0, 2.5)$ weight spatial and temporal urgency. The commanded thrust for the baseline is

\begin{equation}
\mathbf{T}_t = T_{\max}\,\rho_{i^*}\left[ \hat{\mathbf{u}}_{i^*} + \beta\,\hat{\mathbf{v}}_{\text{rel}, i^*} \right], \quad i^* = \arg\max_i \rho_i,
\end{equation}

where $\hat{\mathbf{v}}_{\text{rel}, i}$ is the normalized relative velocity, $\beta = 0.35$ a damping gain, and $T_{\max}$ the thrust authority. 

\paragraph{Impulsive $\Delta v$ Planner} 
Our next baseline is a physically grounded controller that computes the instantaneous relative state $(\vect{r}_{\text{rel}},\, \vect{v}_{\text{rel}})$ between the satellite and the nearest debris and predicts the time to the closest approach

\begin{equation}
t^{*} = -\,\frac{\vect{r}_{\text{rel}} \cdot \vect{v}_{\text{rel}}}{\lVert \vect{v}_{\text{rel}} \rVert^{2}},
\end{equation}

clamped to a finite prediction horizon and estimates the required lateral clearance $d_{\text{req}}$ such that the projected miss distance $\pmiss$ at the TCA exceeds the safety threshold $t_{\text{safe}}$. Then, a constant lateral acceleration is applied as follows:
\begin{equation}
a_{\perp} = \frac{2\,d_{\text{req}}}{t^{*2}}
\end{equation}
along the direction perpendicular to $\vect{v}_{\text{rel}}$ (aligned with the current lateral offset), saturating at the environment’s action bounds and sustaining the burn for $\lceil t^{*}/\Delta t \rceil$ integration steps. This particular implementation was necessary to make the impulsive competitive with the simulator’s discrete-time acceleration model and yield an interpretable avoidance behavior.

\paragraph{Deep Q-Network (DQN)}
\label{sec:DQN_Method}
To evaluate our PPO model we provide a learning-based baseline by implementing a DQN using a discrete-action formulation. The continuous thrust space of the avoidance environment was discretized into a symmetric action set
\begin{equation}
\mathcal{A} = \{\mathbf{0},\, \pm a_0 \hat{\mathbf{e}}_x,\; \pm a_0 \hat{\mathbf{e}}_y,\; \pm a_0 \hat{\mathbf{e}}_z\},
\end{equation}
where $a_0$ is the action magnitude specified in the configuration file and $\hat{\mathbf{e}}_{x,y,z}$ are the body-frame unit vectors. This discretization was applied using a wrapper that maps the agent’s action index to a clipped thrust vector, ensuring compatibility with Stable-Baselines3's DQN implementation.

We train the DQN with a two-layer MLP policy using the following same hyperparameters and reward as our PPO method. We use a smaller architecture of two layers of 64 nodes each as this results in an overall improvement in the policy's performance versus having a matching structure as our PPO policy. Unlike the PPO agent, the DQN does not optimize over a continuous action manifold; instead, it must approximate long-horizon collision avoidance using only this restricted, fixed action set. This limitation is reflected in its high-variance behavior and reduced reliability during evaluation.

\paragraph{No-action controller} Lastly, a no-action baseline serves as a lower bound that coasts without intervention and verifies that the scenario set is non-trivial.

\subsubsection{Deterministic Evaluation Pipeline}
\begin{figure}
    \centering
    \includegraphics[width=0.5\textwidth]{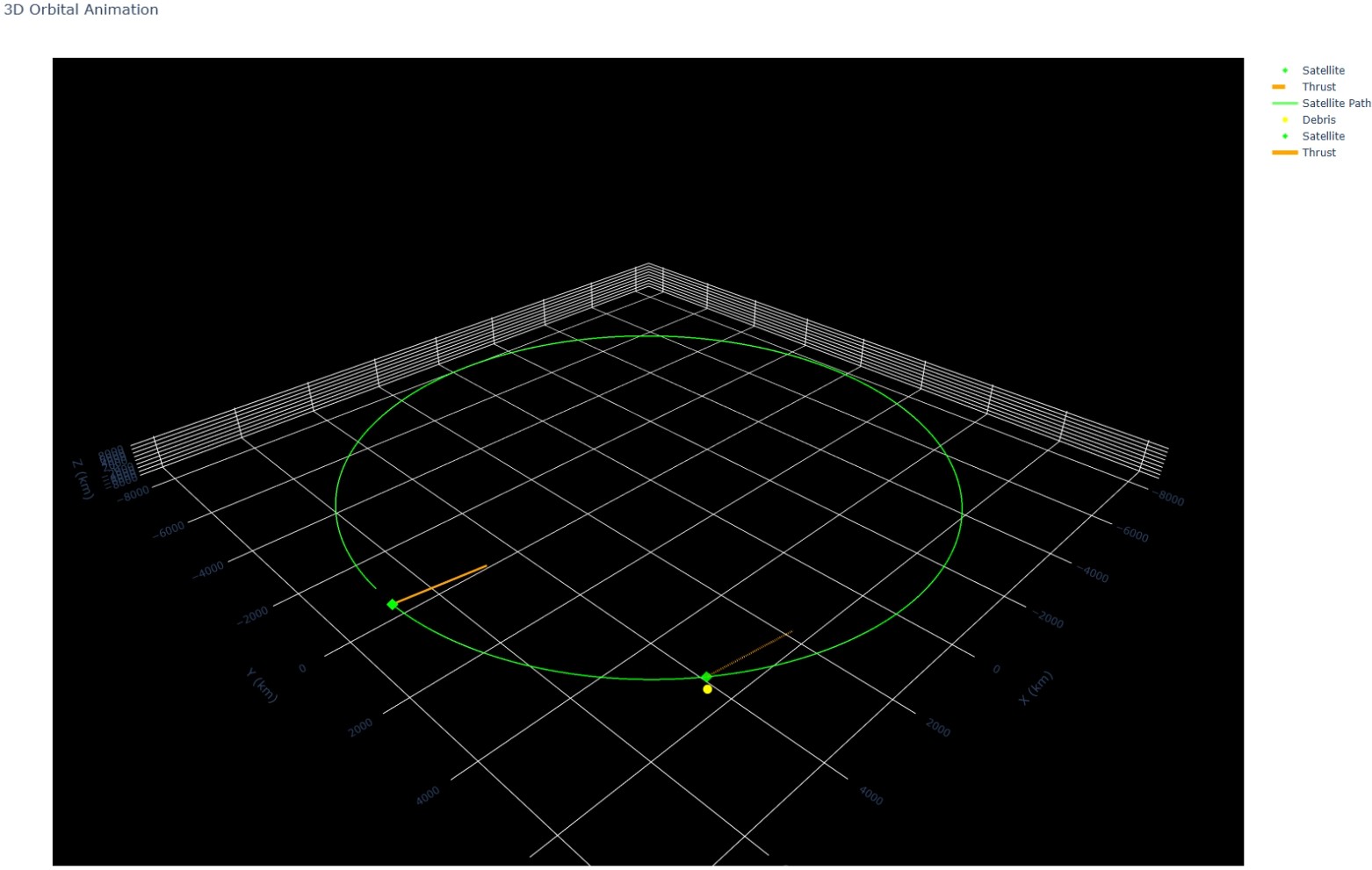}
    \caption{Evaluation Avoidance Sample
    \\ Green Dot: Agent Location
    \\ Green Line: Agent Orbit Trajectory
    \\ Yellow Dot: Debris Location
    \\ Orange Line: Agent Thrust Vector
    }
    \label{fig:Avoidance_visual}
\end{figure}

\begin{figure}
    \centering
    \includegraphics[width=0.5\textwidth]{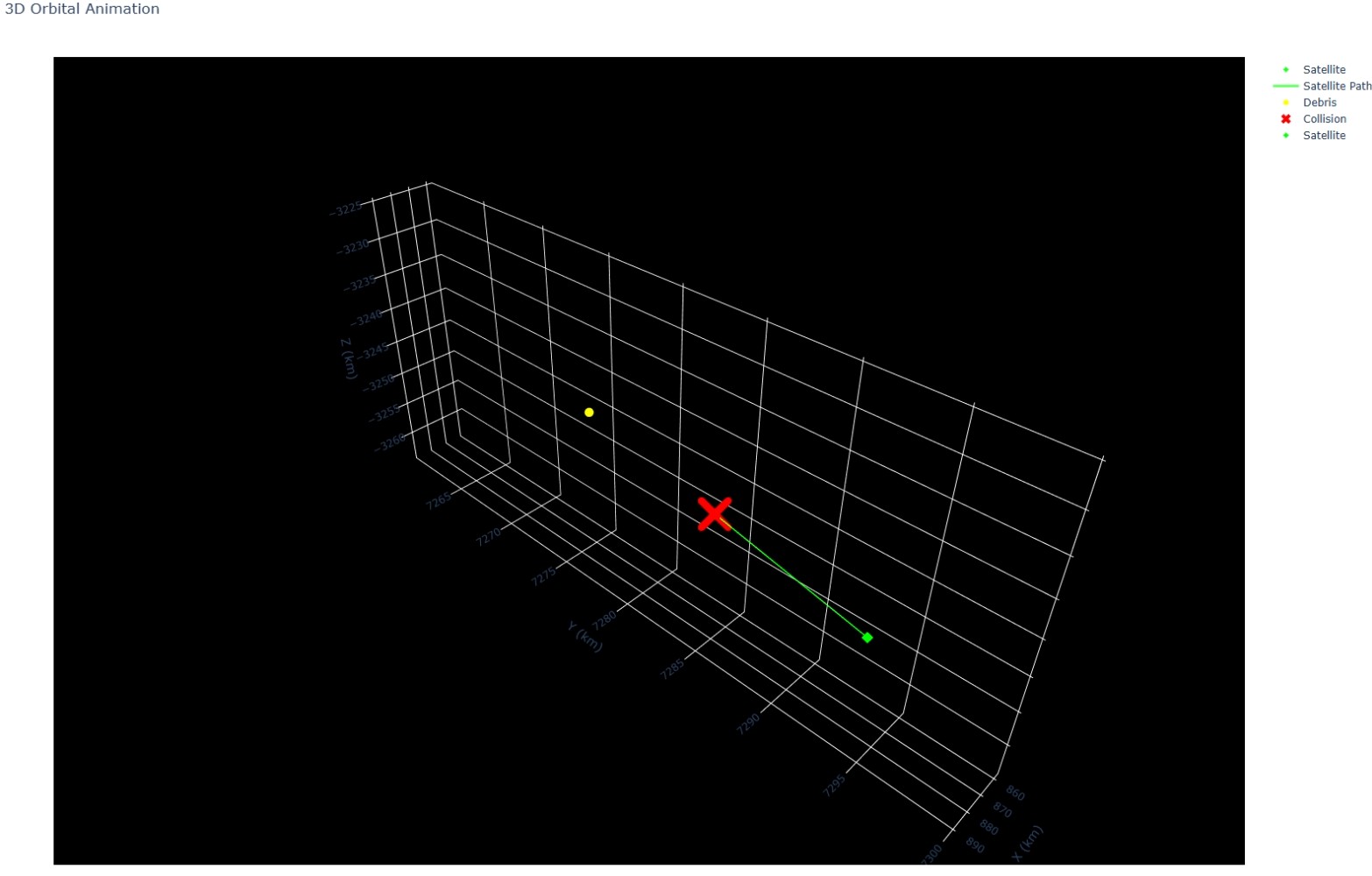}
    \caption{Evaluation Collision Sample
    \\ Green Dot: Agent Location
    \\ Green Line: Agent Orbit Trajectory
    \\ Yellow Dot: Debris Location
    \\ Red Marker: Agent Collision Point
    }
    \label{fig:collision_visual}
\end{figure}

Deterministic evaluation is a large focus in our work; thus, it is possible for others to reproduce our results and potentially expand upon our work. Our evaluation script enforces shared seeds across controllers, re-instantiates environments per policy, and records the following metrics. These metrics include the collision/success rates, episode length, total reward, delta-v usage, minimum approach distance, and fuel consumption. Additionally, per-episode metrics and seeds for direct inspection, and per-step time-series (elapsed time, delta-v increments, cumulative delta-v, and minimum distance). After completing the evaluation, the results are exported to the user's desired path. By saving these results along with the deterministic training of our simulation, we create a fully reproducible system that allows others to continue our work. Samples of our evaluation can be seen in Figs. \ref{fig:Avoidance_visual} and \ref{fig:collision_visual}.

\section{Results}

\begin{figure*}[t]
    \centering
    \begin{subfigure}{0.45\textwidth}
        \centering
        \includegraphics[width=\textwidth]{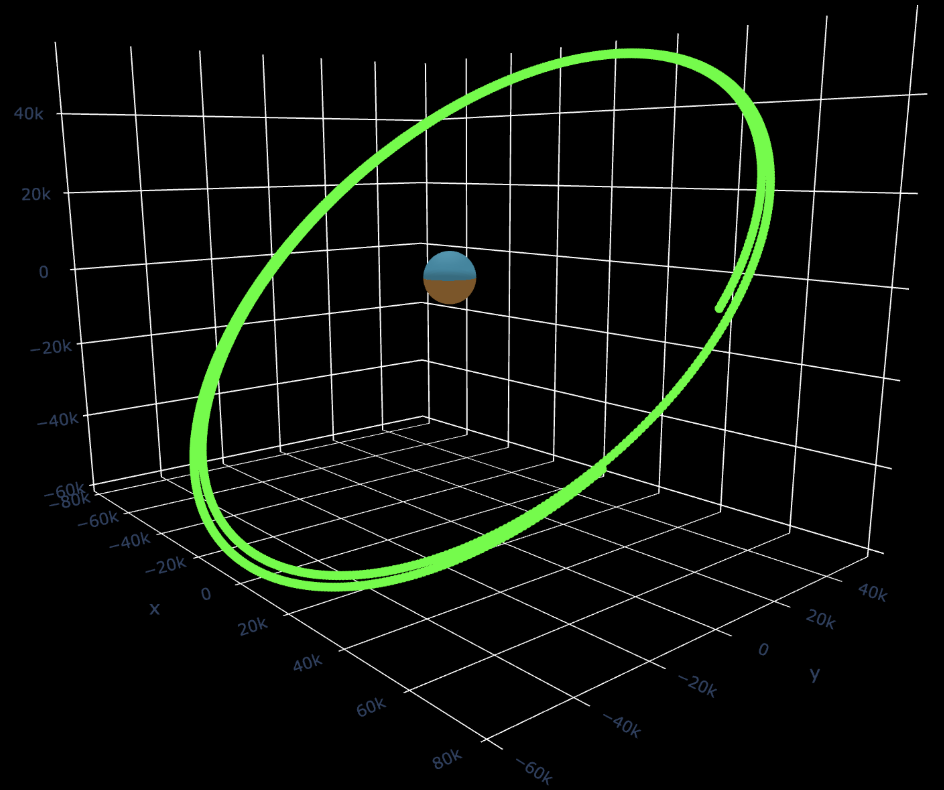}
        \label{fig:result1}
    \end{subfigure}
    \hfill
    \begin{subfigure}{0.45\textwidth}
        \centering
        \includegraphics[width=\textwidth]{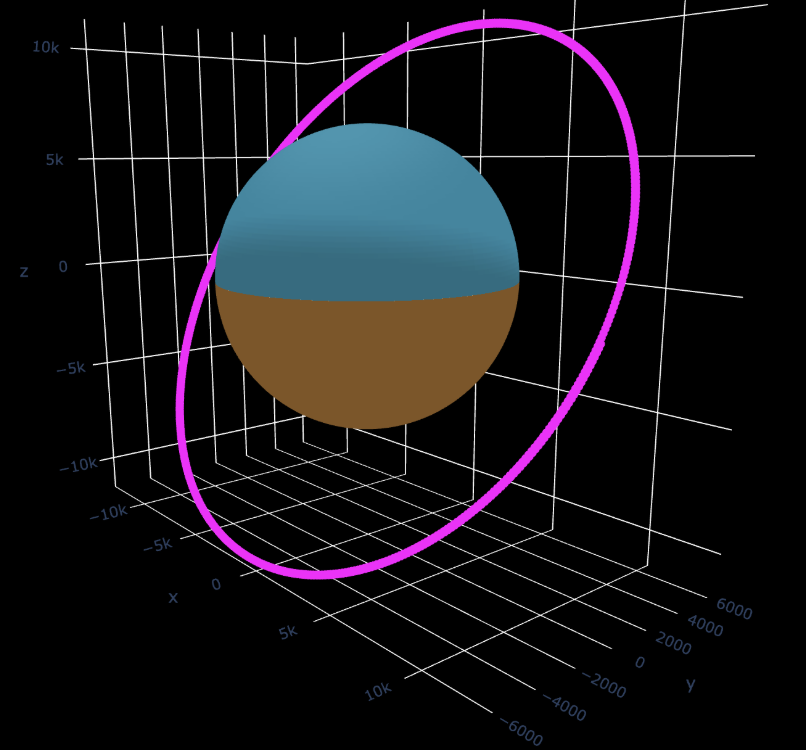}
        \label{fig:result2}
    \end{subfigure}
    
    \caption{Sample Agent Trajectories}
    \label{fig:agent_trajectories}
\end{figure*}

\begin{table*}[t]
\centering
\caption{Agent Performance (1,000-run evaluation)}
\begin{tabular}{@{}lrrrrrrrrr@{}}
\toprule
\textbf{Policy} &
\multicolumn{1}{c}{\textbf{Collision Rate}} &
\multicolumn{1}{c}{\textbf{Collisions}} &
\multicolumn{1}{c}{\textbf{Success Rate}} &
\multicolumn{1}{c}{\textbf{Avg Reward}} &
\multicolumn{1}{c}{\textbf{Avg Steps}} &
\multicolumn{1}{c}{\textbf{Avg $\Delta v$ (m/s)}} &
\multicolumn{1}{c}{\textbf{Avg Min Dist (m)}} &
\multicolumn{1}{c}{\textbf{Fuel Used (kg)}} \\
\midrule
PPO & \textbf{2.50\%} & \textbf{25} & \textbf{97.50\%} & \textbf{3.47$\times$10\textsuperscript{7}} & \textbf{976.20} & \textbf{888.410} & \textbf{1,428.59} & \textbf{254.492} \\
DQN & 62.00\% & 620 & 38.00\% & -4.74$\times10^{4}$ & 319.68 & 147.748 & 176{,}055.46 & 45.915 \\
Rule-based & 79.30\% & 793 & 20.70\% & 7.35$\times$10\textsuperscript{6} & 276.89 & 437.606 & 235.01 & 137.666 \\
Impulsive & 72.50\% & 725 & 27.50\% & 1.00$\times$10\textsuperscript{7} & 321.13 & 12.175 & 269.54 & 4.125 \\
No-action & 100.00\% & 1000 & 0.00\% & 3.87$\times$10\textsuperscript{5} & 63.96 & 0.000 & 140.11 & 0.000 \\
\bottomrule
\end{tabular}
\label{tab:performance}
\end{table*}

\begin{table}
\centering
\caption{Termination counts (1{,}000-run evaluation)}
\begin{tabular}{@{}lrrrr@{}}
\toprule
\textbf{Policy} &
\multicolumn{1}{c}{\textbf{Success}} &
\multicolumn{1}{c}{\textbf{Collision}} &
\multicolumn{1}{c}{\textbf{Fuel Depleted}} &
\multicolumn{1}{c}{\textbf{Timeout}} \\
\midrule
PPO & \textbf{970} & \textbf{25} & \textbf{5} & \textbf{0} \\
DQN & 200 & 620 & 0 & 180 \\
Rule-based & 207 & 793 & 0 & 0 \\
Impulsive & 275 & 725 & 0 & 0 \\
No-action & 0 & 1000 & 0 & 0 \\
\bottomrule
\end{tabular}
\label{tab:termination_counts}
\end{table}

\begin{table*}[t]
\centering
\caption{Mean reward components per episode (1,000-run evaluation)}
\begin{tabular}{@{}lrrrrrrrr@{}}
\toprule
\textbf{Policy} &
\multicolumn{1}{c}{\textbf{Survival}} &
\multicolumn{1}{c}{\textbf{Distance}} &
\multicolumn{1}{c}{\textbf{Closing}} &
\multicolumn{1}{c}{\textbf{Milestone}} &
\multicolumn{1}{c}{\textbf{$\Delta v$ Penalty}} &
\multicolumn{1}{c}{\textbf{Cumulative Penalty}} &
\multicolumn{1}{c}{\textbf{Total}} \\
\midrule
PPO &
\textbf{1.95$\times$10\textsuperscript{3}} &
\textbf{3.48$\times$10\textsuperscript{7}} &
\textbf{5.21} &
\textbf{9.74$\times$10\textsuperscript{1}} &
\textbf{$-8.05\times10^{2}$} &
\textbf{$-5.32\times10^{4}$} &
\textbf{3.47$\times$10\textsuperscript{7}} \\
DQN &
6.38 &
9.86$\times10^{2}$ &
0.31 &
1.12 &
-1.74$\times10^{3}$ &
-4.60$\times10^{4}$ &
-4.74$\times10^{4}$ \\
Rule-based &
5.52$\times$10\textsuperscript{2} &
7.37$\times$10\textsuperscript{6} &
1.19 &
2.18$\times$10\textsuperscript{1} &
$-1.25\times10^{3}$ &
$-1.33\times10^{4}$ &
7.35$\times$10\textsuperscript{6} \\
Impulsive &
6.41$\times$10\textsuperscript{2} &
1.00$\times$10\textsuperscript{7} &
0.61 &
2.75$\times$10\textsuperscript{1} &
$-5.95$ &
0.00 &
1.00$\times$10\textsuperscript{7} \\
No-action &
1.26$\times$10\textsuperscript{2} &
3.87$\times$10\textsuperscript{5} &
0.00 &
0.00$\times$10\textsuperscript{0} &
0.00 &
0.00 &
3.87$\times$10\textsuperscript{5} \\
\bottomrule
\end{tabular}
\label{tab:reward_components}
\end{table*}

\begin{figure}
    \centering
    \includegraphics[width=1.0\linewidth]{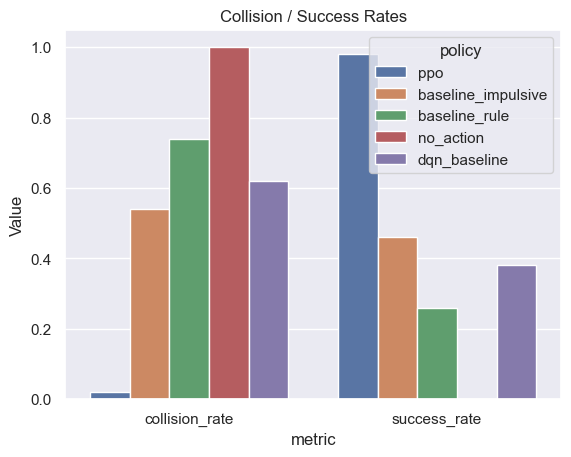}
    \caption{Collision Success Rates}
    \label{fig:collision_success_rates}
\end{figure}

During our deterministic evaluations, we executed 1,000 episodes per policy using the test configuration outlined in Table \ref{tab:scenario_physical}. Table \ref{tab:performance} summarizes the quantitative comparisons. As shown in Table \ref{tab:termination_counts}, the PPO agent succeeds in 975 of 1,000 trials, with the remaining failures attributable to deliberate risk-taking near the boundary. This shows that the agent learned effective collision avoidance strategies, resulting in a 76.8 percentage-point lower collision rate compared to the rule-based baseline policy, and 70.0 percentage-point lower than the impulsive baseline.

\subsection{Policy Behavior}

\subsubsection{PPO Agent}
During training and evaluation, several behaviors led to an increased performance of our agent. First, once the agent receives an observation of the debris' location, it begins a proactive burn. This burn allows it to exploit the thrust limit and widen the miss window, thereby ensuring safe avoidance. Generally, we found that the agent spent approximately the first $150$ seconds increasing the separation to greater than $500$\,m before coasting for the rest of the simulation. This behavior can be observed in Figs. \ref{fig:delta_v_distribution} and \ref{fig:min_distance_distribution_v1}, where the agent’s mean curve quickly exits the danger zone and maintains a 1--2\,km buffer. A notable point that we discuss later is that the agent achieves a higher cumulative $\Delta v$ and fuel consumption, which is consistent with our reward design that tolerates aggressive maneuvers near collisions. Examples of the agents resulting trajectories can be seen in Fig. \ref{fig:agent_trajectories}.

\subsubsection{Rule-Based Policy}
The rule-based policy achieved a 20.7\% success rate, which, while better than the no-action baseline, is still an insufficient score. This likely results from a lack of long-horizon planning, with it often using the majority of its avoidance budget early on, leading to no possible late-stage corrections. Fig. \ref{fig:min_distance_distribution_v1} shows that spacing rarely exceeds $\sim300$\,m, with Fig. \ref{fig:delta_v_distribution} highlighting the discrete $\Delta v$ bursts without sustained follow-through profiles.

\subsubsection{Impulse}
\begin{figure}
    \centering
    \includegraphics[width=1.0\linewidth]{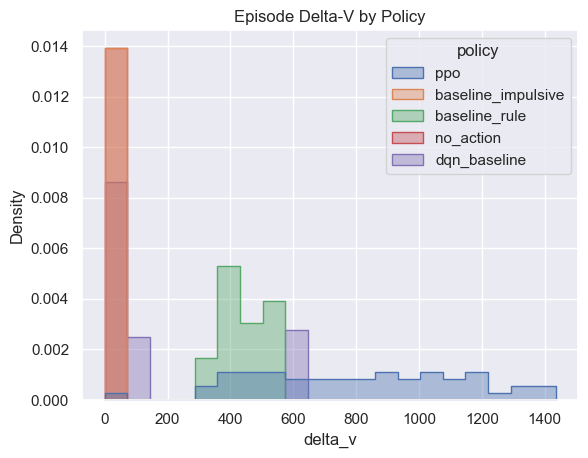}
    \caption{Cumulative Delta-V Distribution}
    \label{fig:delta_v_distribution}
\end{figure}

The impulse baseline displayed a similar behavior to the rule-based baseline, with it's actions and $\Delta v$ spending tapering off very quickly. The baseline began with large burns at the encounter onset rather than the gradual maneuvers, leading to a distance offset initially but not enough for collision avoidance (Figs. \ref{fig:delta_v_distribution}, \ref{fig:min_distance_distribution_pair}). This led to limited trajectory corrections, while it was more accurate and fuel-efficient than the rule-based baseline (27.5\% success rate), it still did not surpass our PPO agent. As a result, it maintains very low cumulative $\Delta v$ (12.2,m/s) and minimal fuel usage (4.1,kg), prioritizing conservation over consistent avoidance. Overall, this displays the limitations of open-loop impulsive avoidance, where while it is fuel-efficient, it fails to ensure safety in dynamic multi-body environments.  

\subsubsection{Deep Q-Network (DQN)}

\begin{figure*}[t]
    \centering
    \begin{subfigure}[t]{0.48\linewidth}
        \centering
        \includegraphics[width=\linewidth]{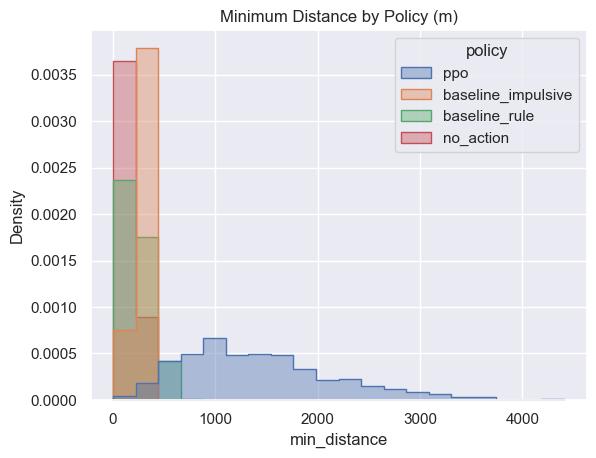}
        \caption{Without DQN}
        \label{fig:min_distance_distribution_v1}
    \end{subfigure}
    \hfill
    \begin{subfigure}[t]{0.48\linewidth}
        \centering
        \includegraphics[width=\linewidth]{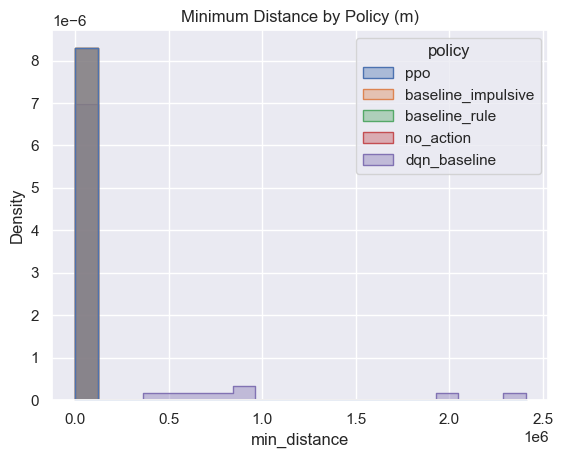}
        \caption{With DQN}
        \label{fig:min_distance_distribution_v2}
    \end{subfigure}
    \caption{Comparison of Minimum Distance Distributions With and Without DQN}
    \label{fig:min_distance_distribution_pair}
\end{figure*}

In the evaluation of DQN, it exhibited high-variance behavior, with its trajectories often diverging into large, unstable orbits. This is what resulted in the disproportionately large average minimum distance when it did successfully avoid, shown in \ref{fig:min_distance_distribution_pair} and \ref{fig:min_distance_distribution_v2}. DQN showed difficulty with long-horizon planning, often failing to apply early corrective burns needed for safe avoidance. Additionally, DQN produced noisy and inconsistent $\Delta v$ usage, with small unstructured thrusts rather than coordinated maneuvers. Many episodes terminated via \texttt{max\_steps} or drifting into wide orbits rather than achieving stable controlled avoidance. One notable point is this policy was much more conservative with its fuel usage in comparison to the other baselines and our own PPO policy. Overall, the policy failed to learn a stable avoidance strategy, highlighting the limitations of DQN in continuous, long-duration orbital tasks and displaying the efficiency of PPO.

\subsubsection{No-Action Policy}
We utilized this as an adversarial baseline to display the result if no action was taken. This shows that only 0.1\% of the trajectories were naturally safe, confirming the difficulty of the encounter set.

\subsection{Fuel Performance}
During the evaluation, the PPO agent consumed 254 kg of fuel compared to the rule-based controller's 138 kg and impulsive's 4 kg, likely due to the reward function penalizing collision more than fuel usage. This is a notable increase, showing a much lesser regard for fuel usage; however, this higher usage is a trade-off for over double the increase in performance, with the PPO’s average reward being nearly five times higher. Therefore, we believe that the increased usage is offset by the increase in performance, with the delta-v efficiency being acceptable for high-priority avoidance maneuvers. 


\begin{figure}
    \centering
    \includegraphics[width=1.0\linewidth]{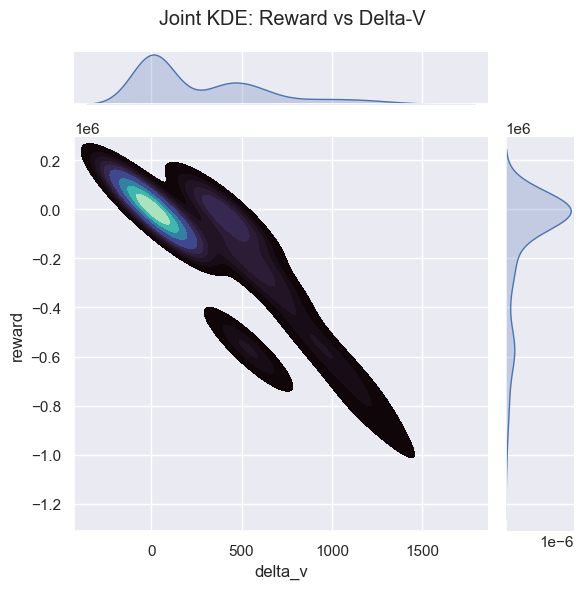}
    \caption{Reward vs.\ Delta-V (Joint Kernel Density Estimation (KDE))}
    \label{fig:reward_vs_delta_v_joint_kde}
\end{figure}

\subsection{Cumulative Rewards}

Table \ref{tab:reward_components} shows the resulting reward from \ref{alg:RewardFunct} into constituent terms. The PPO agent’s dominance stems from strongly positive survival and distance components that outweigh fuel penalties. In contrast, the baselines accrue modest survival reward and pay proportionally smaller penalties, yet fail to offset the collision losses. Fig. \ref{fig:reward_vs_delta_v_joint_kde} shows two manifolds: high reward/high $\Delta$v avoidance (PPO) and low reward/low $\Delta$v failures (baselines). The elongated lobes of the Kernel Density Estimation quantify the trade-off frontier between fuel expenditure and reward.

\section{Limitations}

Despite promising results, several limitations remain:

\subsection{Error Analysis}

As discussed previously, PPO achieves higher success but expends more $\Delta v$ than baselines. This is due to a reward imbalance, with an overemphasis on collision avoidance, leading to suboptimal fuel use. In future work, we will focus on balancing the reward function to improve the fuel consumption while maintaining the performance.

\subsection{Modeling and Evaluation Limits}

Currently, there are several limitations to the implementation of the agent with the simulation environment, preventing it from representing the real world. First, we provide an agent with a perfect sensing of the environment and full noise-free state knowledge. This differs from real missions that require filters for delayed and uncertain observations. We also use a simplified thrust model, ignoring slew rates, burn durations, and control windows, all of which are crucial factors in real $\text{TT\&C}$ operations. For future work, we will address both of these issues, no longer providing the agent with full information on the environment and develop a more representative thrust model.

\subsection{Real-Time Deployment Challenges}

A potential application of our simulation would be the development and deployment of learning-based collision avoidance systems on operational satellites. To further explore this possibility, several real-time constraints would need to be addressed. Firstly, satellite processors often have limited onboard compute, with strict power and memory budgets. Due to this, using a continuous PPO akin to ours would be challenging without model compression or lightweight architectures. The next challenge would include latency and communication windows, with decision making often being delayed due to uplink/downlink schedules. This would require a policy that can operate autonomously with intermittent ground support. Lastly, our currently simulation does not include any realistic safety and verification requirements. Real world policies must be constrained to avoid unsafe actions, often interacting with rule-based systems. Currently our simulation is unconstrained as to best evaluate policy performance, but to explore real-time deployment formal safety guarantees would need to be incorporated.

\section{Future Work}

\begin{figure}
    \centering
    \includegraphics[width=0.45\textwidth]{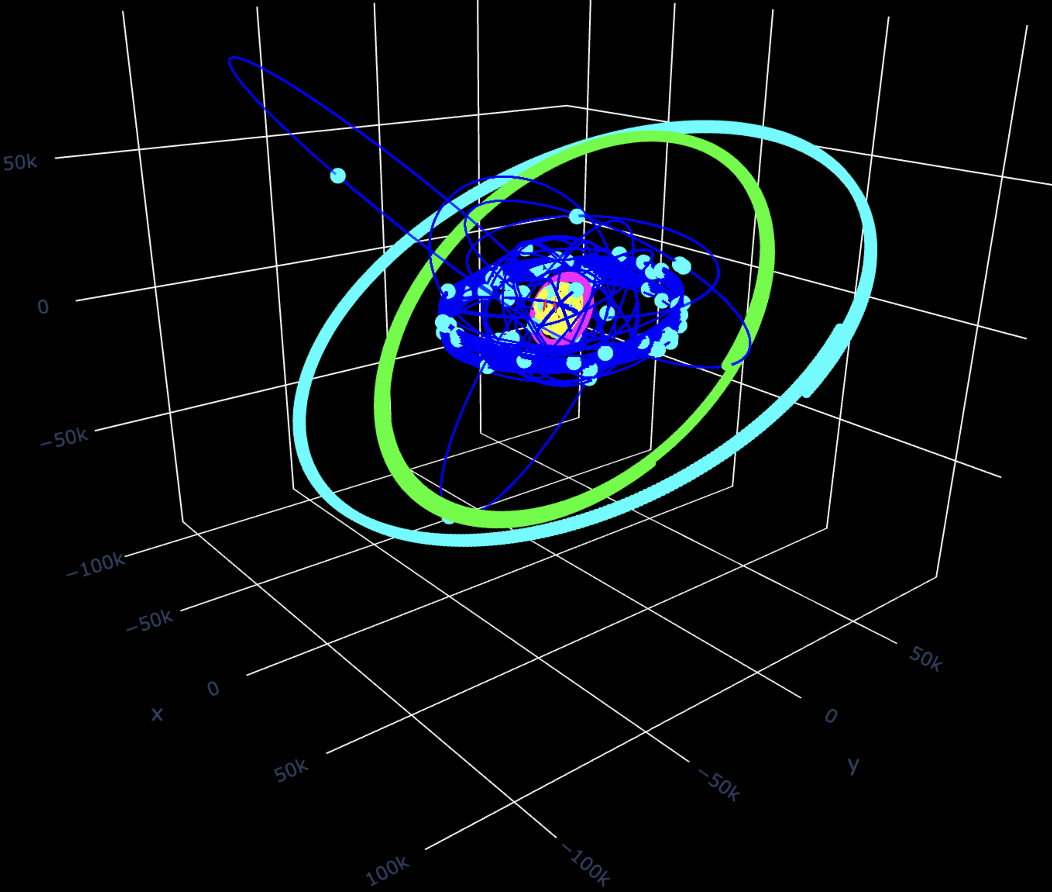}
    \caption{Example swarm simulation scenario.}
    \label{fig:swarm-example}
\end{figure}

In future work, as stated in the limitations, we will focus on implementing realistic sensing, which is representative of the real world. This includes observation noise, delays, and partial observability by using recurrent or belief-based policies. In addition, we intend to add operational constraints often used in spaceflight operations, including enforcing burn magnitude, direction, and timing limits with mission-level cost tradeoffs. We plan to explore hybrid control systems by combining the commonly used rule-based safety logic with learned policies for fallback and safety handovers. Additionally, our work does not feature many modern baselines that are often used owing to initial development constraints, project timelines, and the complexity of their implementation. Our hope was that the current baselines, although older, still provide the necessary insight into the improvements that our agent provides in collision avoidance. Therefore, we intend to address this in future work by implementing advanced guidance (e.g., MPC, optimal control, conjunction analysis) for stronger comparisons. Finally, we will explore the implementation of swarm modeling in our environment, extending to multi-agent avoidance and debris-field interactions (Fig.~\ref{fig:swarm-example}).

\section{Conclusion}

Our study introduces a publicly available collision avoidance simulation environment for real-time mapping of space debris and active satellites in orbit, along with the capabilities of training and evaluating autonomous agents. Our pipeline focuses on reproducible results, including deterministic training and evaluation of methods, along with telemetry data logging for transparent peer review and rapid experimentation, including set-scenario blueprints, per-episode logs, summary tables, and evaluation seeds. We found that this level of packaging and transparency is uncommon in orbital RL research, thus lowering the barrier for comparative studies.

Additionally, we create an autonomous Proximal Policy Optimization agent that demonstrates a collision avoidance capability that exceeds that of classical methods and other RL policies, with the agent’s distributed burn strategy and small repeated thrusts achieving smoother and more consistent avoidance than the baseline’s sporadic spike. We also integrate a curriculum learning schedule and reward shaping oriented towards collision avoidance that balances collision penalties with fuel-conservation incentives, enabling the responsible use of thrust authority. For a competitive evaluation, we compare our model to industry baselines, a risk-aware heuristic controller that reacts to distance and time-to-collision, and an impulsive strategy grounded in classical orbital mechanics and conjunction analysis. This comparison validates the advantage of our method, owing to its increased performance over these baselines. Overall, our framework establishes a robust foundation for training and evaluating RL-based collision avoidance in contested orbital environments while providing a new competitive agent.

\appendices

\section{Nomenclature}

\begin{table}[H]
\caption{Nomenclature}
\label{tab:nomenclature}
\setlength{\tabcolsep}{3pt}
\renewcommand{\arraystretch}{1.2}
\begin{tabular}{|p{32pt}|p{90pt}|p{115pt}|}
\hline
\textbf{Symbol} & 
\textbf{Quantity} & 
\textbf{Description / Units} \\
\hline
$\vect{r}$ & Position vector & Satellite position in ECI frame [m] \\
$\vect{v}$ & Velocity vector & Satellite velocity in ECI frame [m/s] \\
$\dot{\vect{r}}$ & Velocity derivative & $\dot{\vect{r}}=\vect{v}$ \\
$\dot{\vect{v}}$ & Acceleration vector & Total acceleration incl. gravity and thrust [m/s$^2$] \\
$\mu_{\oplus}$ & Earth gravitational parameter & $3.986\times10^{14}$ [m$^3$/s$^2$] \\
$\mu_k$ & Perturber parameter & Grav. parameter of Moon/Sun [m$^3$/s$^2$] \\
$\vect{r}_k$ & Perturber position & Moon/Sun position relative to Earth [m] \\
$m_t$ & Satellite mass & Includes structure + remaining fuel [kg] \\
$\vect{T}_t$ & Thrust vector & Applied thrust at time $t$ [N] \\
$T_{\max}$ & Max thrust & Engine thrust limit [N] \\
$v_e$ & Exhaust velocity & $v_e = I_{sp} g_0$ [m/s] \\
$I_{sp}$ & Specific impulse & Engine efficiency [s] \\
$g_0$ & Standard gravity & $9.80665$ [m/s$^2$] \\
$\Delta v_t$ & Velocity increment & Speed change over $\Delta t$ [m/s] \\
$\Delta t$ & Time step & Integration interval [s] \\
$d_t$ & Separation distance & $\min_i \|\vect{r}_{sat}-\vect{r}_i\|$ [m] \\
$d_{\text{coll}}$ & Collision threshold & Unsafe separation distance [m] \\
$d_{\text{safe}}$ & Safety buffer radius & $d_{\text{coll}} + B$ [m] \\
$r_{\text{eff}}$ & Effective safety radius & $r_{sat}+r_{deb}+r_{margin}$ [m] \\
$\pmiss$ & Projected miss distance & Estimated closest approach [m] \\
$v_{\text{close},t}$ & Relative closing speed & Instantaneous LOS approach rate [m/s] \\
$\rho_i$ & Hazard score & Risk weight for debris $i$ \\
$\sigma(\cdot)$ & Sigmoid & Logistic function for $\rho_i$ \\
$\alpha_d,\alpha_\tau$ & Weighting scalars & Distance/TTC scaling parameters \\
$d_i$ & Distance to debris $i$ & $\|r_{sat}-r_i\|$ [m] \\
$\tau_i$ & TTC & Time to closest approach of debris $i$ \\
$\tau_{\text{crit}}$ & Critical TTC & Safety-conditioned threshold [s] \\
$i^*$ & Most threatening index & $i^*=\arg\max_i \rho_i$ \\
$\hat{\mathbf{u}}_i$ & Avoidance direction & Unit vector away from debris $i$ \\
$\hat{\mathbf{v}}_{\text{rel},i}$ & Rel. velocity dir. & Normalized relative velocity \\
$\beta$ & Blending factor & Velocity-alignment weight \\
$\vect{r}_{\text{rel}}$ & Relative position & $(r_{sat}-r_i)$ [m] \\
$\vect{v}_{\text{rel}}$ & Relative velocity & $(v_{sat}-v_i)$ [m/s] \\
$t^*$ & Time of closest approach & $-(r_{\text{rel}}\cdot v_{\text{rel}})/\|v_{\text{rel}}\|^2$ \\
$a_{\perp}$ & Required normal accel. & $2d_{req}/t^{*2}$ [m/s$^2$] \\
$d_{\text{req}}$ & Required clearance & Target minimum pass distance [m] \\
$\mathcal{A}$ & Action set & $\{0,\pm a_0\hat{e}_x,\pm a_0\hat{e}_y,\pm a_0\hat{e}_z\}$ \\
$a_0$ & Action magnitude & Discrete thrust level [N] \\
$\Delta a$ & Action change & $\|a_t - a_{t-1}\|$ \\
$C_{\Delta v}$ & $\Delta v$ cost term & Context-aware burn penalty \\
$\lambda_{\text{surv}}$, $\lambda_{\text{coast}}$, $\lambda_{\text{smooth}}$, $\lambda_{\text{proj}}$ &
Reward coefficients & Survival, coasting, smoothness, projection shaping \\
$c_d$, $c_1$, $c_2$, $c_{\Sigma v}$ &
Cost constants & Distance, $\Delta v$, cumulative-$\Delta v$ shaping \\
$p_{\text{large}}$ &
Large burn penalty & Quadratic penalty on large impulses \\
$v_{\text{soft}}$ &
Soft $\Delta v$ cap & Cumulative burn threshold [m/s] \\
$R_{\text{mile}}$ &
Milestone reward & Bonus for long-term survival \\
$R_t$ &
Step reward & Reward at timestep $t$ \\
$R_{\text{total}}$ &
Episode reward & Sum of step rewards \\
$P_{\text{coll}}$ &
Collision penalty & Large negative terminal reward \\
\hline
\multicolumn{3}{p{250pt}}{\textit{Indicator function: }$\mathbf{1}[\cdot]=1$ if condition true, else $0$. All quantities use SI units unless stated otherwise.}\\
\hline
\end{tabular}
\end{table}

\section{Training and Evaluation Configurations}
\label{configurations}

\begin{table}[H]
\centering
\caption{Training reward shaping coefficients}
\begin{tabular}{@{}lr@{}}
\toprule
\textbf{Parameter} & \textbf{Value} \\
\midrule
\texttt{collision\_penalty}      & 1000.0 \\
\texttt{survival\_reward}        & 0.02 \\
\texttt{distance\_shaping\_coeff}& 0.0002 \\
\texttt{coast\_bonus}            & 0.05 \\
\texttt{delta\_v\_linear\_cost}  & 8.0 \\
\texttt{delta\_v\_quadratic\_cost}& 0.8 \\
\texttt{large\_burn\_penalty}    & 4.0 \\
\texttt{smoothness\_penalty}     & 0.008 \\
\texttt{cumulative\_dv\_penalty} & 0.3 \\
\texttt{projected\_miss\_reward} & 0.0002 \\
\texttt{milestone\_reward}       & 0.4 \\
\bottomrule
\end{tabular}
\label{tab:train_reward_values}
\end{table}

\begin{table}[H]
\centering
\caption{Evaluation reward shaping coefficients}
\begin{tabular}{@{}lr@{}}
\toprule
\textbf{Parameter} & \textbf{Value} \\
\midrule
\texttt{collision\_penalty}      & 50.0 \\
\texttt{survival\_reward}        & 2.0 \\
\texttt{distance\_shaping\_coeff}& 1.0 \\
\texttt{coast\_bonus}            & 3.0 \\
\texttt{delta\_v\_linear\_cost}  & 0.5 \\
\texttt{delta\_v\_quadratic\_cost}& 0.05 \\
\texttt{large\_burn\_penalty}    & 0.5 \\
\texttt{smoothness\_penalty}     & 0.0005 \\
\texttt{cumulative\_dv\_penalty} & 0.02 \\
\texttt{projected\_miss\_reward} & 2.0 \\
\texttt{milestone\_reward}       & 10.0 \\
\bottomrule
\end{tabular}
\label{tab:eval_reward_values}
\end{table}

\begin{table}[H]
\centering
\caption{Training runtime settings}
\begin{tabular}{@{}lr@{}}
\toprule
\textbf{Parameter} & \textbf{Value} \\
\midrule
\texttt{vec\_env}     & \texttt{dummy} \\
\texttt{num\_envs}    & 8 \\
\texttt{device}       & \texttt{cpu} \\
\texttt{total\_timesteps} & 1{,}000{,}000 \\
\texttt{eval\_freq}        & 20{,}000 \\
\bottomrule
\end{tabular}
\label{tab:train_runtime}
\end{table}

\begin{table}[H]
\centering
\caption{Evaluation scenario and parameters}
\begin{tabular}{@{}lr@{}}
\toprule
\textbf{Parameter} & \textbf{Value} \\
\midrule
\texttt{collision.guarantee} & \texttt{false} \\
\texttt{collision.probability} & 1.0 \\
\texttt{collision.ttc\_steps} & 20 \\
\texttt{debris\_radius\_m}     & 50.0 \\
\texttt{max\_thrust}           & 0.15 \\
\bottomrule
\end{tabular}
\label{tab:scenario_physical}
\end{table}

\begin{table}[H]
\centering
\caption{PPO hyperparameters}
\begin{tabular}{@{}lr@{}}
\toprule
\textbf{Parameter} & \textbf{Value} \\
\midrule
\texttt{n\_steps}      & 2048 \\
\texttt{batch\_size}   & 256 \\
\texttt{n\_epochs}     & 10 \\
\texttt{learning\_rate}& 0.0003 \\
\texttt{lr\_schedule}  & \texttt{constant} \\
\texttt{ent\_coef}     & 0.01 \\
\texttt{vf\_coef}      & 0.5 \\
\texttt{max\_grad\_norm} & 0.5 \\
\texttt{gamma}         & 0.995 \\
\texttt{gae\_lambda}   & 0.95 \\
\texttt{clip\_range}   & 0.2 \\
\texttt{target\_kl}    & 0.01 \\
\texttt{net\_arch}     & [256, 256, 128] \\
\bottomrule
\end{tabular}
\label{tab:ppo_hparams}
\end{table}

\bibliographystyle{ieeetr} 
\bibliography{main}

@misc{esa2024,
  author       = {{European Space Agency (ESA)}},
  title        = {{The Current State of Space Debris}},
  year         = {2024},
  howpublished = {\url{https://www.esa.int/Space_Safety/Space_Debris/The_current_state_of_space_debris}},
}

@misc{esa2021numbers,
  author       = {{ESA Space Debris Office}},
  title        = {{Space Debris by the Numbers}},
  year         = {2021},
  howpublished = {\url{https://www.esa.int/Safety_Security/Space_Debris/Space_debris_by_the_numbers}},
}

@article{Kessler1978,
  author        = {Kessler, Donald J. and Cour-Palais, Burton G.},
  title         = {{Collision Frequency of Artificial Satellites: The Creation of a Debris Belt}},
  journal       = {Journal of Geophysical Research},
  volume        = {83},
  number        = {A6},
  pages         = {2637--2646},
  year          = {1978},
  doi           = {10.1029/JA083iA06p02637}
}

@misc{PayloadSpace2024,
  author        = {{PayloadSpace News}},
  title         = {{ESA Report Shows Unsustainable Levels of Orbital Debris}},
  year          = {2024},
  howpublished  = {\url{https://payloadspace.com/esa-report-shows-unsustainable-levels-of-orbital-debris}},
}

@misc{FAA2023,
  author       = {{Federal Aviation Administration}},
  title        = {FAA Proposed Rule Would Reduce Growth of Debris from Commercial Space Vehicles},
  year         = {2023},
  howpublished = {\url{https://www.faa.gov/newsroom/faa-proposed-rule-would-reduce-growth-debris-commercial-space-vehicles}},
}

@misc{UNU2024b,
  author       = {{Arcep}},
  title        = {Satellites and the Environment: When the Promises of Megaconstellations Collide with the Limits of Space},
  year         = {2021},
  howpublished = {\url{https://en.arcep.fr/news/press-releases/view/n/satellites-and-environnement-event-ademe-arcep-cnes-201123.html}},
}

@article{SpaceIntelDebris2024,
  author       = {de Selding, Peter B.},
  title        = {{ESA's annual debris report: Fragmentations averaging 10 per year; surprising debris effect on Earth observation imagery}},
  journal      = {Space Intel Report},
  year         = {2025},
  month        = apr,
  howpublished = {\url{https://www.spaceintelreport.com/esas-annual-debris-report-fragmentations-averaging-10-per-year-surprising-debris-effect-on-earth-observation-imagery/}},
}

@inproceedings{Johnson2010OrbitalDebris,
  author       = {Johnson, Nicholas L.},
  title        = {Orbital Debris: The Growing Threat to Space Operations},
  booktitle    = {Proceedings of the 33rd Annual Guidance and Control Conference},
  year         = {2010},
  address      = {Breckenridge, CO, USA},
  organization = {American Astronautical Society},
  url          = {https://ntrs.nasa.gov/citations/20100004498}
}

@article{Tarran2021PrepareForImpact,
  author  = {Tarran, Brian},
  title   = {Prepare for Impact: Space Debris and Statistics},
  journal = {Significance},
  volume  = {18},
  number  = {3},
  pages   = {18--23},
  year    = {2021},
  doi     = {10.1111/1740-9713.01527},
  url     = {https://academic.oup.com/jrssig/article/18/3/18/7038543},
}

@techreport{GAO2023,
  author        = {{U.S. Government Accountability Office}},
  title         = {Satellite Deployments: Expected Growth and Regulatory Implications},
  institution   = {U.S. Government Accountability Office},
  number        = {GAO-23-105678},
  year          = {2023},
  howpublished  = {\url{https://www.gao.gov/products/gao-23-105678}},
}

@misc{UNU2024c,
  author       = {{Union of Concerned Scientists}},
  title        = {UCS Satellite Database},
  year         = {2024},
  howpublished = {\url{https://www.ucsusa.org/resources/satellite-database}},
}

@misc{ESA2025CDM,
  author       = {{European Space Agency}},
  title        = {Reentry and Collision Avoidance},
  howpublished = {\url{https://www.esa.int/Space_Safety/Space_Debris/Reentry_and_collision_avoidance}},
}

@techreport{ScienceDirectCAMThreshold,
  author       = {{European Space Agency}},
  title        = {{ESA Space Debris Mitigation Requirements}},
  institution  = {European Space Agency},
  number       = {ESSB-ST-U-007 Issue 1},
  type         = {Standard},
  address      = {Paris, France},
  year         = {2023},
  howpublished = {\url{https://technology.esa.int/upload/media/ESA-Space-Debris-Mitigation-Requirements-ESSB-ST-U-007-Issue1.pdf}},
}

@article{NatureThreshold,
  author  = {Boley, Aaron C. and Byers, Michael},
  title   = {Satellite mega-constellations create risks in Low Earth Orbit, the atmosphere and on Earth},
  journal = {Scientific Reports},
  volume  = {11},
  number  = {10642},
  year    = {2021},
  doi     = {10.1038/s41598-021-89909-7},
  url     = {https://www.nature.com/articles/s41598-021-89909-7}
}

@misc{SpacePolicyOnline2019,
  author       = {{European Space Agency}},
  title        = {ESA Spacecraft Dodges Large Constellation},
  year         = {2019},
  howpublished = {\url{https://www.esa.int/Space_Safety/ESA_spacecraft_dodges_large_constellation}}}

@misc{ArXivCAMPlanning,
  author       = {Thompson, Amy},
  title        = {Traffic Jams From Satellite Fleets Are Imminent{\textemdash}What It Means for Earth},
  year         = {2019},
  howpublished = {\url{https://observer.com/2019/09/satellite-space-congestion-spacex-starlink-esa-aeolus/}}}

@misc{SpaceStackExchangeISS,
  author        = {Howell, Elizabeth},
  title         = {How Often Does the International Space Station Have to Dodge Space Debris?},
  year          = {2023},
  howpublished  = {\url{https://www.space.com/international-space-station-space-dodge-debris-how-often}}}

@misc{ArcivAutonomyStrain2024,
  author       = {{New Space Economy Editorial Team}},
  title        = {How Do Satellites Avoid Collisions?},
  year         = {2025},
  howpublished = {\url{https://newspaceeconomy.ca/2025/11/10/how-do-satellites-avoid-collisions/}}}

@misc{SpaceComFuelImpact2024,
  author       = {Pultarova, Tereza},
  title        = {How many satellites can we safely fit in Earth orbit?},
  year         = {2023},
  howpublished = {\url{https://www.space.com/how-many-satellites-fit-safely-earth-orbit}}
}

@article{NatureCounterManeuver2024,
  author  = {Rossi, Alessandro and S{\'a}nchez-Ortiz, Noelia and David, Emmanuelle and Opromolla, Roberto and Grishko, Dmitriy},
  title   = {Future Activities in the Near-Earth Space in the Face of Ever-Increasing Space Traffic},
  journal = {Acta Astronautica},
  volume  = {225},
  pages   = {891--897},
  year    = {2024},
  doi     = {10.1016/j.actaastro.2024.09.063},
  url     = {https://doi.org/10.1016/j.actaastro.2024.09.063}
}

@article{NatureTransparency2024,
  author  = {Blount, P. J.},
  title   = {Space Traffic Management: Standardizing On-Orbit Behavior},
  journal = {AJIL Unbound},
  volume  = {113},
  pages   = {120--124},
  year    = {2019},
  doi     = {10.1017/aju.2019.17},
  url     = {https://www.cambridge.org/core/journals/american-journal-of-international-law/article/space-traffic-management-standardizing-onorbit-behavior/336B7F6141E7F2174013FAE508B9AACD}
}

@book{Alpaydin2021,
  author    = {Alpaydin, Ethem},
  title     = {Introduction to Machine Learning},
  year      = {2020},
  edition   = {4},
  publisher = {MIT Press},
  address   = {Cambridge, MA, USA},
}

@incollection{alloghani2020,
  author    = {Alloghani, Mohamed and Al-Jumeily, Dhiya and Mustafina, Jamila and Hussain, Abir and Aljaaf, Ahmed J.},
  title     = {A Systematic Review on Supervised and Unsupervised Machine Learning Algorithms for Data Science},
  booktitle = {Supervised and Unsupervised Learning for Data Science},
  pages     = {3--21},
  year      = {2020},
  publisher = {Springer},
  address   = {Cham, Switzerland},
  doi       = {10.1007/978-3-030-22475-2_1},
}

@article{franccois2018,
  title     = {An Introduction to Deep Reinforcement Learning},
  author    = {Fran{\c{c}}ois-Lavet, Vincent and Henderson, Peter and Islam, Riashat and Bellemare, Marc G. and Pineau, Joelle},
  journal   = {Foundations and Trends{\textregistered} in Machine Learning},
  volume    = {11},
  number    = {3--4},
  pages     = {219--354},
  year      = {2018},
  publisher = {Now Publishers Inc.}
}

@article{schulman2017ppo,
  author  = {Schulman, John and Wolski, Filip and Dhariwal, Prafulla and Radford, Alec and Klimov, Oleg},
  title   = {Proximal Policy Optimization Algorithms},
  journal = {arXiv preprint arXiv:1707.06347},
  year    = {2017},
  url     = {https://arxiv.org/abs/1707.06347}
}

@book{chan2008spacecraft,
  author    = {Chan, K.},
  title     = {Spacecraft Collision Probability},
  publisher = {Aerospace Press},
  address   = {El Segundo, CA, USA},
  year      = {2008}
}

@book{battin1999astrodynamics,
  author    = {Battin, R. H.},
  title     = {An Introduction to the Mathematics and Methods of Astrodynamics},
  publisher = {AIAA},
  address   = {Reston, VA, USA},
  year      = {1999}
}

@techreport{nasa2015cara,
  author       = {Newman, Lauri K.},
  title        = {NASA Conjunction Assessment Risk Analysis (CARA) Program: Updated Requirements Architecture},
  institution  = {NASA Goddard Space Flight Center},
  address      = {Greenbelt, MD, USA},
  year         = {2019},
  number       = {GSFC-E-DAA-TN39817},
  howpublished = {\url{https://ntrs.nasa.gov/citations/20190001474}},
}

@article{esa2021sdo,
  author  = {Klinkrad, H. and Beltrami, P. and Krag, H. and others},
  title   = {The {ESA} Space Debris Mitigation Handbook 2002},
  journal = {Advances in Space Research},
  volume  = {34},
  number  = {5},
  pages   = {1251--1259},
  year    = {2004},
  doi     = {10.1016/j.asr.2003.01.018},
}

@article{park2019decision,
  title   = {Risk Assessment of Recent High-Interest Conjunctions},
  author  = {Alfano, Sal and Oltrogge, Daniel L. and Krag, Holger and Merz, Klaus and Hall, Robert},
  journal = {Acta Astronautica},
  volume  = {184},
  pages   = {241--250},
  year    = {2021},
  doi     = {10.1016/j.actaastro.2021.04.009},
  url     = {https://www.sciencedirect.com/science/article/pii/S0094576521001545}
}

@article{hwang2020heuristic,
  title   = {Assessing and Minimizing Collisions in Satellite Mega-Constellations},
  author  = {Reiland, Nathan and Rosengren, Aaron J. and Malhotra, Renu and Bombardelli, Claudio},
  journal = {Advances in Space Research},
  volume  = {67},
  number  = {11},
  pages   = {3755--3774},
  year    = {2021},
  doi     = {10.1016/j.asr.2021.01.010},
  url     = {https://doi.org/10.1016/j.asr.2021.01.010}
}

@article{mnih2015dqn,
  title         = {Human-level control through deep reinforcement learning},
  author        = {Mnih, Volodymyr and Kavukcuoglu, Koray and Silver, David
                   and Rusu, Andrei A. and Veness, Joel and Bellemare, Marc G.
                   and Graves, Alex and Riedmiller, Martin and Fidjeland, Andreas K.
                   and Ostrovski, Georg and Petersen, Stig and Beattie, Charles
                   and Sadik, Amir and Antonoglou, Ioannis and King, Helen
                   and Kumaran, Dharshan and Wierstra, Daan and Legg, Shane
                   and Hassabis, Demis},
  journal       = {Nature},
  volume        = {518},
  number        = {7540},
  pages         = {529--533},
  year          = {2015},
  publisher     = {Nature Publishing Group},
  doi           = {10.1038/nature14236},
  url           = {https://doi.org/10.1038/nature14236}
}

@misc{ESAAuto2024,
  author       = {{European Space Agency}},
  title        = {Automating Collision Avoidance},
  year         = {2019},
  howpublished = {\url{https://www.esa.int/Space_Safety/Space_Debris/Automating_collision_avoidance}},
}

@misc{SPD3_2018,
  author       = {{The White House}},
  title        = {Space Policy Directive-3: National Space Traffic Management Policy},
  year         = {2018},
  howpublished = {\url{https://trumpwhitehouse.archives.gov/presidential-actions/space-policy-directive-3-national-space-traffic-management-policy/}},
}

@misc{AI4EarthScienceCDMTransparency,
  author       = {Uriot, Thomas and Izzo, Dario and Martinez-Heras, Jos{\'e} A. and Letizia, Francesca and Siminski, Jan and Merz, Klaus},
  title        = {Collision Avoidance Challenge Dataset},
  year         = {2021},
  howpublished = {\url{https://doi.org/10.5281/zenodo.4463683}},
}

@inproceedings{KesslerLibrary2021,
  author    = {Acciarini, Giacomo and Pinto, Francesco and Letizia, Francesca and Martinez-Heras, Jos{\'e} A. and Merz, Klaus and Bridges, Christopher and Baydin, At{\i}l{\i}m G{\"u}ne{\c{s}}},
  title     = {Kessler: A Machine Learning Library for Spacecraft Collision Avoidance},
  booktitle = {Proceedings of the 8th European Conference on Space Debris},
  year      = {2021},
  address   = {Darmstadt, Germany},
  publisher = {ESA Space Debris Office},
  url       = {https://strathprints.strath.ac.uk/76217/}
}

@inproceedings{KesslerLibrary2022,
  author    = {Acciarini, Giacomo and Baresi, Nicola and Bridges, Christopher and Felicetti, Leonard and Hobbs, Stephen and Baydin, At{\i}l{\i}m G{\"u}ne{\c{s}}},
  title     = {Observation Strategies and Megaconstellations Impact on Current LEO Population},
  booktitle = {2nd NEO and Debris Detection Conference},
  year      = {2023},
  address   = {Darmstadt, Germany}
}

@techreport{NASA_CARA2019,
  author       = {Krage, Frederic J.},
  title        = {NASA Spacecraft Conjunction Assessment and Collision Avoidance Best Practices Handbook},
  institution  = {National Aeronautics and Space Administration (NASA)},
  number       = {NASA/SP-20230002470},
  year         = {2023},
  address      = {Washington, DC, USA},
  url          = {https://ntrs.nasa.gov/citations/20230002470},
}

@inproceedings{POMDP2023,
  author       = {Abdin, Adam and Bourriez, Nicolas and Loizeau, Adrien},
  title        = {Spacecraft Autonomous Decision-Planning for Collision Avoidance: a Reinforcement Learning-Based Approach},
  booktitle    = {Proceedings of the 74th International Astronautical Congress (IAC-23), Symposium A6.7: Operations in Space Debris Environment},
  year         = {2023},
  address      = {Baku, Azerbaijan},
  organization = {International Astronautical Federation},
  url          = {https://iafastro.directory/iac/archive/browse/IAC-23/A6/7/76800/},
}

@inproceedings{Smith2024,
  author    = {Solomon, Alexandru and P{\u{a}}duraru, Ciprian},
  title     = {Collision Avoidance and Return Manoeuvre Optimisation for Low-Thrust Satellites Using Reinforcement Learning},
  booktitle = {Proceedings of the 17th International Conference on Agents and Artificial Intelligence (ICAART 2025)},
  year      = {2025},
  volume    = {3},
  pages     = {1009--1016},
  publisher = {SCITEPRESS},
  doi       = {10.5220/0013249000003890},
  url       = {https://www.scitepress.org/Papers/2025/132490/132490.pdf},
}

@article{Kazemi2024,
  author       = {Mu, Chaoxu and Liu, Shuo and Lu, Ming and Liu, Zhaoyang and others},
  title        = {Autonomous Spacecraft Collision Avoidance with a Variable Number of Space Debris Based on Safe Reinforcement Learning},
  journal      = {Aerospace Science and Technology},
  volume       = {149},
  pages        = {109131},
  year         = {2024},
  doi          = {10.1016/j.ast.2024.109131},
}

@article{poliastro2019,
  title={{Poliastro}: An Open Source Python Library for Orbital Mechanics},
  author={Benito, J. L. and Lara, M. and Cappelletti, A. and others},
  journal={Journal of Open Source Software},
  volume={4},
  number={43},
  pages={1884},
  year={2019},
  doi={10.21105/joss.01884}
}

@article{astropy2022,
  title={The Astropy Project: Sustaining and Growing a Community-Oriented Open-Source Project and the Latest Major Release (v5.0)},
  author={Astropy Collaboration and Price-Whelan, A. M. and Lim, P. L. and Earl, N. and Starkman, N. and others},
  journal={The Astrophysical Journal},
  volume={935},
  number={2},
  pages={167},
  year={2022},
  doi={10.3847/1538-4357/ac7c74}
}

@misc{CelesTrakTLE2024,
  author       = {{CelesTrak}},
  title        = {NORAD Two-Line Element Sets for Satellite Tracking},
  year         = {2024},
  howpublished = {\url{https://celestrak.org/NORAD/elements/index.php?FORMAT=csv}},
}

@misc{OpenAI_Gym2016,
  author        = {Brockman, G. et al.},
  title         = {OpenAI Gym: A toolkit for developing and comparing RL algorithms},
  year          = {2016},
  howpublished  = {\url{https://arxiv.org/abs/1606.01540}},
}

@misc{StableBaselines2017,
  author        = {Dhariwal, P. et al.},
  title         = {Stable Baselines: High-Quality Implementations of RL Algorithms},
  year          = {2017},
  howpublished  = {\url{https://github.com/openai/baselines}},
}

@misc{BGR2024,
  author       = {Pine, Damien},
  title        = {It's time to clean up space junk before orbits become `unusable,' according to new ESA report},
  year         = {2025},
  howpublished = {\url{https://www.livescience.com/space/its-time-to-clean-up-space-junk-before-orbits-become-unusable-according-to-new-esa-report}},
}

@techreport{UNU2024,
  author      = {Eberle, Caitlyn and Sebesvari, Zita},
  title       = {Technical Report: Space Debris},
  institution = {United Nations University},
  address     = {Bonn, Germany},
  year        = {2022},
  url         = {https://digitallibrary.un.org/record/4068239},
}

@article{ESAChallenge2021,
  author  = {Uriot, Thomas and Izzo, Dario and Sim{\~o}es, Lu{\'\i}s and Abay, Rukiye and Einecke, Nicole and Rebhan, Stefan and Martinez-Heras, Jos{\'e} A. and Letizia, Francesca and Siminski, Jan and Merz, Klaus},
  title   = {Spacecraft collision avoidance challenge: Design and results of a machine learning competition},
  journal = {Astrodynamics},
  year    = {2022},
  volume  = {6},
  number  = {2},
  pages   = {121--140},
  doi     = {10.1007/s42064-021-0101-5},
}

@inproceedings{ardaens2013collision,
  title={Collision avoidance maneuver planning using a multi-objective optimization approach},
  author={Ardaens, J.-S. and Früh, C.},
  booktitle={2013 6th International Conference on Recent Advances in Space Technologies (RAST)},
  pages={603--608},
  year={2013},
  organization={IEEE},
  doi={10.1109/RAST.2013.6581273}
}

@article{armellin2021collision,
  author        = {R. Armellin},
  journal       = {Acta Astronautica},
  title         = {Collision Avoidance Maneuver Optimization with a Multiple-Impulse Convex Formulation},
  year          = {2021},
  volume        = {186},
  pages         = {347--362},
  doi           = {10.1016/j.actaastro.2021.05.046}
}

\begin{IEEEbiography}[{\includegraphics[width=1in,height=1.25in,clip,keepaspectratio]{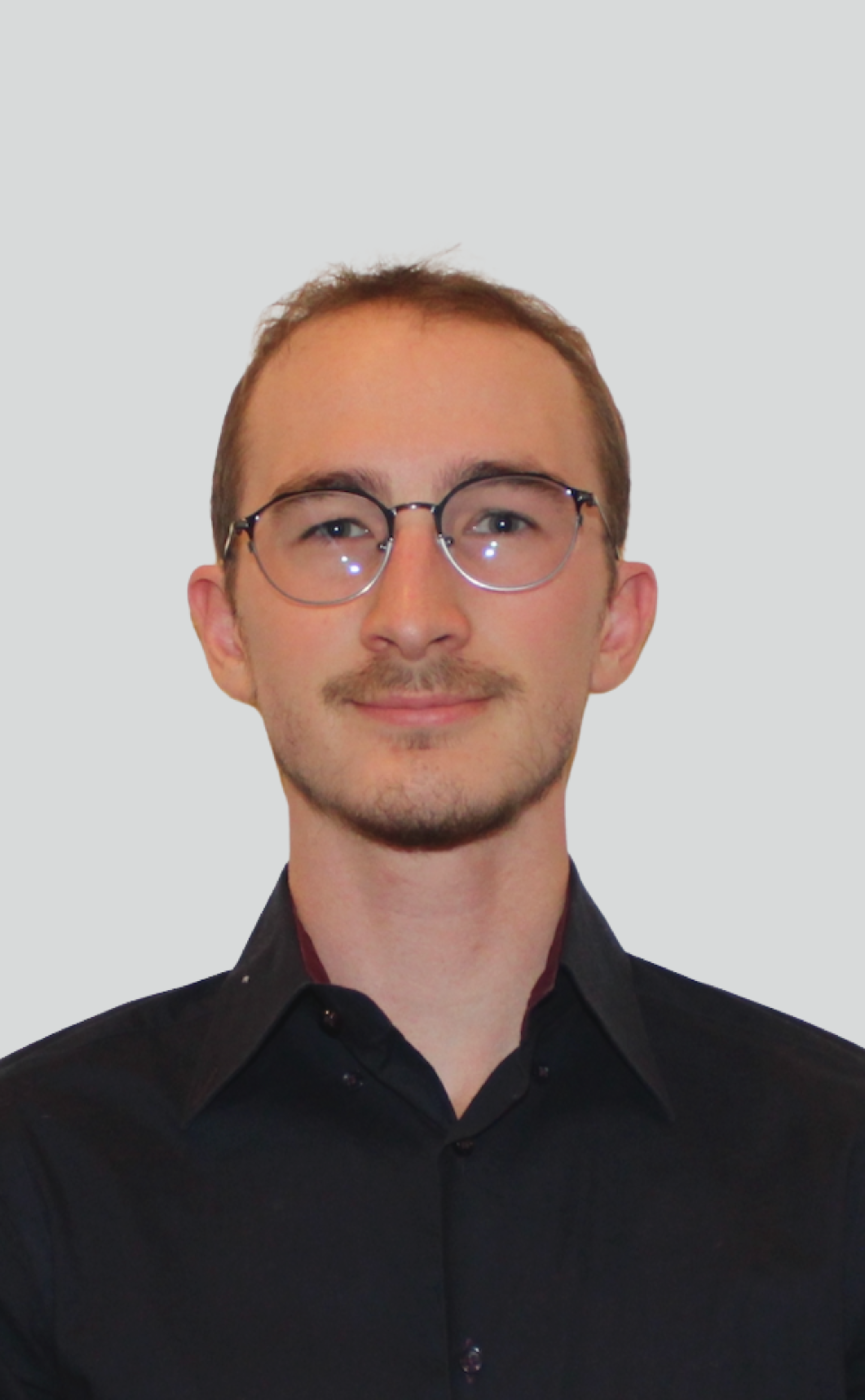}}]{Logan Luna}
(Member, IEEE) received the B.S. degree in computer science with a minor in computational mathematics from Embry-Riddle Aeronautical University, Daytona Beach, FL, USA. He is currently pursuing the M.S. degree in computer science with the School of Computer Science, College of Computing, Georgia Institute of Technology, Atlanta, GA, USA. He is a Department of Defense SMART Scholar. His research focuses on machine learning, reinforcement learning, and autonomous systems, with applications in orbital collision avoidance, intrusion detection, and sensor fusion. He was a recipient of the IEEE-Eta Kappa Nu Best Paper Award and the 2025 Outstanding Undergraduate in Computer Science Award.
\end{IEEEbiography}

\begin{IEEEbiography}[{\includegraphics[width=1in,height=1.25in,clip,keepaspectratio]{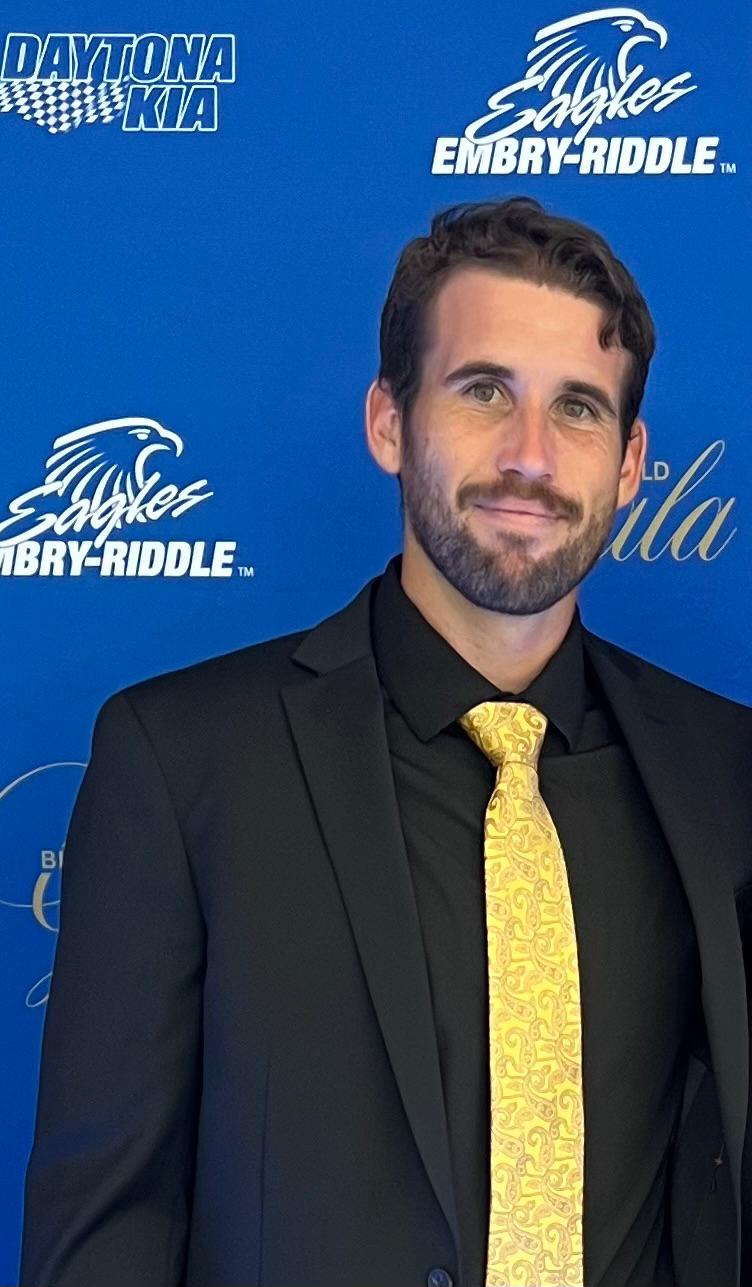}}]{Juan Ortiz Couder}
(Member, IEEE) received the B.S. and M.S. degrees in software engineering from Embry-Riddle Aeronautical University, where he is currently pursuing the Ph.D. degree. His research centers on the integration of machine learning into the cybersecurity field from a software engineering perspective. In addition, his research interest includes software engineering education, with a particular focus on incorporating current AI trends into academia.
\end{IEEEbiography}

\begin{IEEEbiography}[{\includegraphics[width=1in,height=1.25in,clip,keepaspectratio]{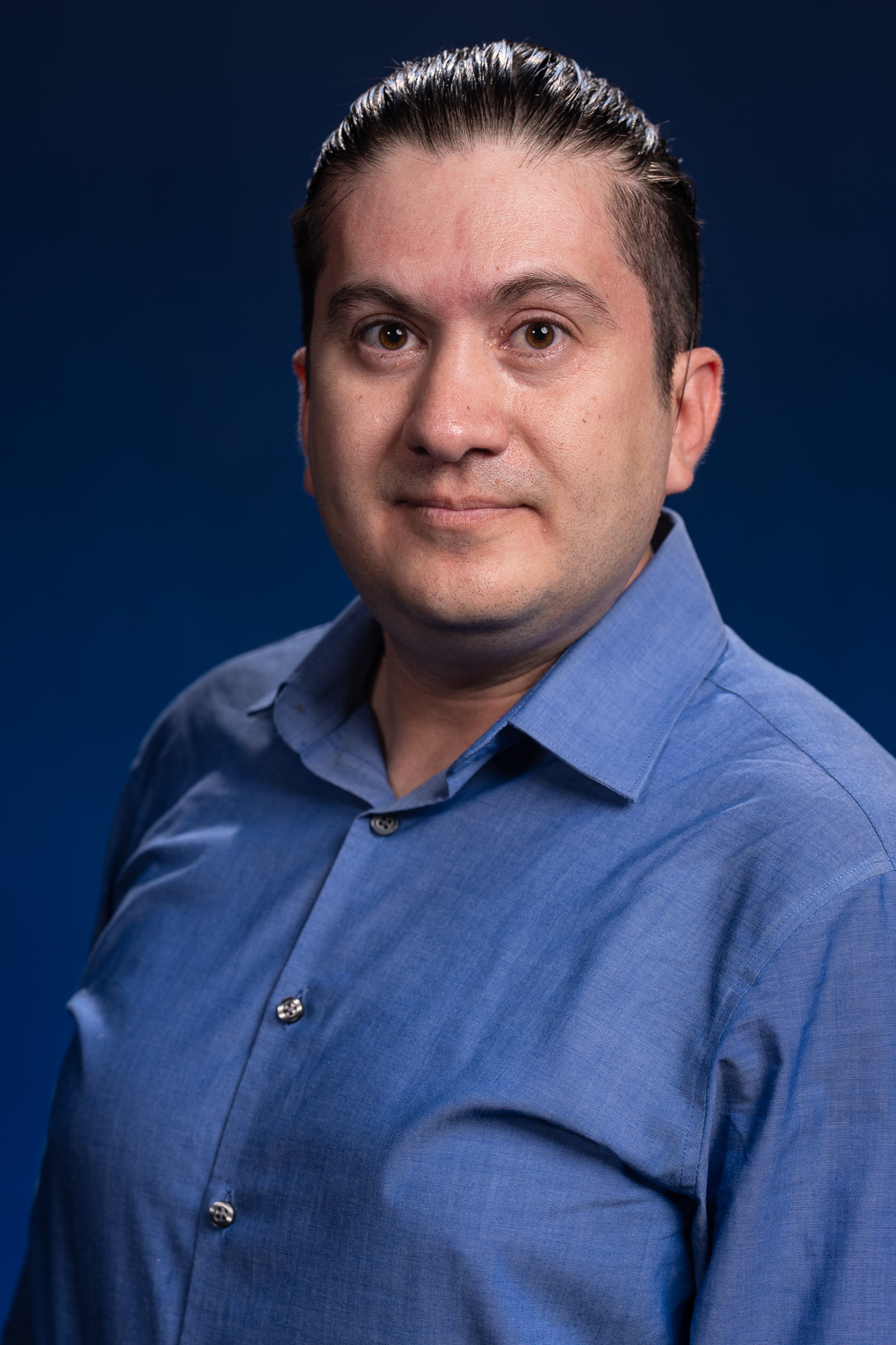}}]{Raul Alejandro Vargas-Acosta}
(Member, IEEE) received the Ph.D. degree in computer science from The University of Texas at El Paso, El Paso, TX, USA, in 2024. He is currently an Assistant Professor with the Department of Electrical Engineering and Computer Science, Embry-Riddle Aeronautical University, Daytona Beach, FL, USA. His research interests include data analytics, knowledge representation, and their applications across diverse domains. He is a member of the Upsilon Pi Epsilon.
\end{IEEEbiography}

\EOD

\end{document}